\documentclass[preprint,review,12pt]{elsarticle}

\usepackage{natbib}
\usepackage{mathtools}
\usepackage{amsmath}
\usepackage{array}
\usepackage{mathrsfs}
\usepackage{amssymb}
\usepackage{booktabs}
\usepackage{algorithm}
\usepackage{algpseudocode}
\usepackage{multirow}
\usepackage{tabularx}
\usepackage{tabulary}
\usepackage{dcolumn}
\usepackage{threeparttable}
\usepackage{array}
\usepackage{bbm}
\usepackage{color, soul}

\usepackage{setspace}
\usepackage{url}
\usepackage{rotating}

\journal{Neurocomputing}

\begin{document}

\begin{frontmatter}



\title{A Composite Quantile Fourier Neural Network \\for Multi-Step Probabilistic Forecasting\\of Nonstationary Univariate Time Series\tnoteref{label1}}


\author{Kostas~Hatalis\corref{cor1}}
\ead{kmh511@lehigh.edu}

\author{Shalinee~Kishore}
\address{Department of Electrical and Computer Engineering, \\Lehigh University, Bethlehem, PA 18105, USA}

\tnotetext[label1]{The research reported in this manuscript was supported by the National Science Foundation; grant number 1442858}
\cortext[cor1]{Corresponding author}

\begin{abstract}
	Point forecasting of univariate time series is a challenging problem with extensive work having been conducted. However, nonparametric probabilistic forecasting of time series, such as in the form of quantiles or prediction intervals is an even more challenging problem. In an effort to expand the possible forecasting paradigms we devise and explore an extrapolation-based approach that has not been applied before for probabilistic forecasting. We present a novel quantile Fourier neural network is for nonparametric probabilistic forecasting of univariate time series. Multi-step predictions are provided in the form of composite quantiles using time as the only input to the model. This effectively is a form of extrapolation based nonlinear quantile regression applied for forecasting. Experiments are conducted on eight real world datasets that demonstrate a variety of periodic and aperiodic patterns. Nine naive and advanced methods are used as benchmarks including quantile regression neural network, support vector quantile regression, SARIMA, and exponential smoothing. The obtained empirical results validate the effectiveness of the proposed method in providing high quality and accurate probabilistic predictions. 
\end{abstract}

\begin{keyword}
probabilistic forecasting \sep
artificial neural networks \sep
time series \sep
multi-step prediction \sep
quantile regression
\end{keyword}

\end{frontmatter}


\section{Introduction}

Univariate time series based deterministic or point forecasting is a well-studied field that has numerous usages. Examples of applications include finance \cite{xue2017financial}, topic behavior \cite{hu2016predicting}, traffic flow \cite{zhou2017delta}, and renewable power \cite{yan2016time}. There are several approaches to forecasting with different classes of methods. Approaches include having a sliding window of past data to predict future data, recurrent models, and extrapolation based regression such as signal approximation \cite{godfrey2017neural}. In all the approaches, methods are divided into two classes as linear or nonlinear. In the first class, methods include linear regression, AutoRegression (AR), AutoRegressive–Moving-Average (ARMA) models, and exponential smoothing. The second class of methods is nonlinear models which are predominantly machine learning based such as support vector regression, nonlinear autoregression neural networks, and recurrent neural networks. A thorough overview of time series and machine learning based deterministic forecasting can be found in \cite{box2015time,langkvist2014review}.

Despite the wide use of deterministic forecasting, it does have a significant disadvantage in that it can result in individual observation errors which can be significant. Additionally, deterministic forecasting lacks information on associated uncertainty. A solution to these problems is Probabilistic Forecasting (PF) where the goal is to produce fully probabilistic predictions that derive quantitative information on the uncertainty. A PF takes the form of a predictive probability distribution over future time horizons and aims to maximize the sharpness of predictive densities while subject to reliability. Sharpness refers to the concentration of the predictive distribution and reliability refers to the accuracy of the forecasted probability in conveying the actual probability of events.

A popular application of PF is in the fields of renewable energies and power systems. A probabilistic forecast is vital for different operations to renewable energy farms. This includes managing the optimal level of generation reserves \cite{doherty2005new}, optimizing production \cite{castronuovo2004optimization}, and bidding strategies for electricity markets \cite{pinson2007trading}. Applications to the power grid include load analysis \cite{yang2017power}, smart meters \cite{taieb2016forecasting}, scheduling \cite{su2014stochastic}, system planning \cite{hong2014long}, unit commitment \cite{botterud2013demand}, and energy trading \cite{andrade2017probabilistic}. A thorough overview of probabilistic wind and solar power forecasting is provided in \cite{zhang2014review} and \cite{van2017review}. Further popularity of PF in energy systems can be seen by the large number of submissions in the Global Energy Forecasting Competition of 2014 \cite{hong2016probabilistic}.

Additional motivation for this work comes from the lack of studies conducted for nonparametric PF of univariate time series, particularly for multi-step prediction. Most nonparametric PF work utilize a multivariate regression based framework, while most univariate time series PFs are made using parametric assumptions, for instance Gaussian intervals. There is such little understanding of prediction intervals for time series forecasting, that the popular M4 Competition for time series forecasting \cite{makridakis2018m4} included a PF track for producing prediction interval forecasts specifically to study their performance in more detail. There own findings demonstrate that standard forecasting methods fail to estimate the uncertainty in time series forecasts properly, and therefore that more research is required in time series PF.

There are several main classes in the type of PF models which include if they are parametric or nonparametric, direct or indirect, and the type of inputs they use for prediction. In PF, we are first trying to predict one of two types of density functions, either parametric or nonparametric. When the future density function is assumed to take a specific distribution, such as the Normal distribution, then this is known as parametric probabilistic forecasting. For processes where no assumption is made about the shape of the distribution, a nonparametric probabilistic forecast can be made. Nonparametric predictions can be made in the form of quantiles, prediction intervals, or full density functions. For example, nonlinear and non-stationary data, such as wind speeds or stocks, may not correspond to fixed or known distributions. When in need of forecasting such data it can be more beneficial to apply a nonparametric probabilistic forecast to estimate the distribution instead then assume it is shaped.

The second classification of PF models is whether they are combined with point forecasts or not. For instance in \cite{huang2017deterministic} a deterministic and probabilistic forecast for wind power is combined. This approach is known as an indirect PF method. First, a point forecast is made such as with support vector regression, and then prediction intervals for point forecast values are obtained with a PF method such as quantile regression. On the other hand, when a PF method estimates future quantiles or prediction intervals without using as input point forecasts, this is known as direct forecasting. The last distinction to be made with PF models is if past lagged data are used as inputs to the forecast model or if future exogenous variables are used too. For instance in renewable power PF, if Numerical Weather Predictions (NWP) exist for each forecasting horizon that we are interested in, then those exogenous NWPs are used as inputs to provide a PF in that prediction horizon. When NWPs are not given then lagged past time values of renewable power can be used for prediction. 

We introduce a new approach to developing a nonparametric direct PF where the input to the model is neither exogenous variables nor past data but instead treats the series as a signal. This approach is motivated by Fourier extrapolation which is the process by which a Fourier transform is applied to a data set to decompose it into a sum of sinusoidal components thus interpreting it as a signal. In time series analysis this is related to Harmonic regression. In accounting for periodic and non-periodic aspects of a signal such as a trend, Fourier Neural Networks (FNN) have been proposed. 

FNNs are feedforward neural networks with sinusoidal activation functions that model the Fourier transform. While other machine learning methods such as convolutional or recurrent neural networks have shown success in time series forecasting \cite{yang2015deep}, these architectures are not well suited for modeling Fourier extrapolation. Thus, most recently, a new FNN called Neural Decomposition (ND) was proposed in \cite{godfrey2017neural} that can decompose a signal into a sum of its constituent parts, model trend, and reconstruct a signal beyond the training samples. ND can provide a prediction by having time as its only input similarly to an inverse Fourier transform. We propose a PF model motivated by the ND model. 

Several works have explored Fourier extrapolation based deterministic forecasting with sinusoidal neural networks \cite{gashler2016modeling,godfrey2017neural,germec2009fourier}, but none have yet explored it for probabilistic forecasting. We are the first to introduce an FNN for forecasting composite quantiles that we dub the quantile Fourier neural network. The main contributions of our approach can be summarized as follows:

\begin{enumerate}
	\item We show how to estimate composite quantiles using a Fourier neural network using a smoothed pinball loss function.
	\item We demonstrate how this extrapolation based quantile forecasting is able to model periodic and non-periodic components of nonstationary time series.
	\item We demonstrate how to conduct multi-step probabilistic forecasts in the form of quantiles or prediction intervals.
	\item We design experiments to validate our approach for direct probabilistic forecasting and provide insight how this method can generalize modeling uncertainty on real-world datasets.
\end{enumerate}

The contents of the paper are: in Section \ref{s:prob} we provide the mathematical background on probabilistic forecasting, quantile regression, and evaluation methods. In Section \ref{s:mod} we review existing architectures of FNNs, go over our model, its architecture, training, and weighting initialization scheme. Results and discussion of validating our method are presented in Section \ref{s:res}. We conclude the paper and review future research directions in Section \ref{s:con}.

\section{Probabilistic Forecasting} \label{s:prob}

This section highlights the underlying mathematics in probabilistic forecasting, overviews linear quantile regression which forms the foundation of our proposed method, and summarizes the main evaluation metric for density forecasts. Given a random variable $Y_t$ at time $ t $, its probability density function is defined as $f_t$ and its the cumulative distribution function as $F_t$. If $ F_{t} $ is strictly increasing, the quantile $ q_{t}^{(\tau)} $ at time $t$ of the random variable $ Y_{t} $ with nominal proportion $ \tau $ is uniquely defined on the value $ x $ such that $ P(Y_t < x) = \tau $ or equivalently as the inverse of the distribution function $ q_{t}^{(\tau)} = F_{t}^{-1}(\tau) $. A quantile forecast $ \hat{q}_{t+z}^{(\tau)} $ is an estimate of the true quantile $ q_{t+z}^{(\tau)} $ for the lead time $ t+z $, given a predictor values. 

Prediction intervals are another type of probabilistic forecast and give a range of possible values within which an observed value is expected to lie with a certain probability $ \beta \in [0,1] $. A prediction interval $ \hat{I}^{(\beta)}_{t+z} $ produced at time $ t $ for future horizon $ t + z $ is defined by its lower and upper bounds, which are the quantile forecasts $ \hat{I}^{(\beta)}_{t+z} = \left[ \hat{q}_{t+z}^{(\tau_{l})} ,\hat{q}_{t+z}^{(\tau_{u})} \right] = \left[ l_{t}^{(\beta)},u_{t}^{(\beta)} \right] $ whose nominal proportions $ \tau_l $ and $ \tau_u $ are such that $ \tau_u - \tau_l = 1-\beta $. A nonparametric probabilistic forecast $ \hat{f}_{t+z} $ \cite{pinson2007non} can be made of the density function by gathering a set of $ M $ quantiles forecasts such that $ \hat{f}_{t+z} = \left\lbrace  \hat{q}_{t+z}^{(\tau_{m})} ,m=1,...,M|0\leq \tau_1 < ... < \tau_M \leq 1 \right\rbrace $ with chosen nominal proportions spread on the unit interval.

Quantile regression is a popular approach for nonparametric probabilistic forecasting. Koenker and Bassett \cite{koenker1978regression} introduce it for estimating conditional quantiles and is closely related to models for the conditional median \cite{koenker2005quantile}. Minimizing the mean absolute function leads to an estimate of the conditional median of a prediction. By applying asymmetric weights to errors through a tilted transformation of the absolute value function, we can compute the conditional quantiles of a predictive distribution. The selected transformation function is the pinball loss function as defined by
\begin{equation} \label{pinball}
\rho_{\tau}(u) = \left\lbrace 
\begin{array}{cl}
\tau u    & \mbox{if } u \geq 0 \\
(\tau-1)u & \mbox{if } u < 0
\end{array} \right.,
\end{equation}

\noindent where $ 0 < \tau < 1 $ is the tilting parameter. The pinball loss function penalizes low-probability quantiles more for overestimation than for underestimation and vice versa in the case of high-probability quantiles. Given a vector of predictors $ X_t $ where $ t = 1,...,N $, a vector of weights $ W $ and intercept $ b $ coefficient in a linear regression fashion, the conditional $ \tau_i $ quantile, for $i \in [0,1]$, is given by $ \hat{q}_{t}^{(\tau_i )} = W^\top X_t +b $. To determine estimates for the weights and intercept for composite quantile regression we solve the following minimization problem
\begin{equation} \label{eq3}
\min_{W,b} \frac{1}{NM} \sum_{i=1}^{M} \sum_{t=1}^{N} \rho_{\tau_{i}} (y_t-\hat{q}_{t}^{(\tau_i)}),
\end{equation}

\noindent where $ y_t $ is the observed value of the predictand and $M$ is the number of quantiles we are estimating. The formulation above in Eq. (\ref{eq3}) can be minimized by a linear program. Here are many variations of QR. In \cite{bremnes2004probabilistic} local QR is applied to estimate different quantiles, while in \cite{nielsen2006using} a spline-based QR is used to estimate quantiles of wind power. In \cite{landry2016probabilistic} quantile loss gradient boosted machines are used to estimate many quantiles, and in \cite{juban2016multiple} multiple quantile regression is used to predict a full distribution with optimization achieved by using the alternating direction method of multipliers. Quantile regression forests \cite{juban2008uncertainty} are another approach in forecasting which are an extension of regression forests based on classification and regression trees. Due to their flexibility in modeling elaborate nonlinear data sets, artificial neural networks are another dominant class of machine learning algorithms that can be used to enhance QR. Taylor \cite{taylor2000quantile} implemented and demonstrated a quantile regression neural network (QRNN) method, combining the advantages of both QR and a neural network. In \cite{xu2016quantile} an autoregressive version of QRNN is used for applications to evaluating value at risk, and \cite{cannon2011quantile} implements the QRNN model in R as a statistical package. We extend the QRNN model to form our quantile Fourier neural network.

\subsection{Evaluation Methods}

In probabilistic forecasting it is essential to evaluate the quantile estimates and derived predictive intervals. To evaluate quantile predictions, one can use the pinball function directly for evaluation called the quantile score (QS). We choose QS as our main quantile measure for the following reasons. When averaged across many quantiles it can evaluate full predictive densities, it is found to be a proper scoring rule \cite{grushka2016quantile}, and it is related to the continuous rank probability score. QS calculated overall $ N $ test observations and $ M $ quantiles is defined as

\begin{equation*}
QS = \sum_{t=1}^{N} \sum_{m=1}^{M} \rho_{\tau_m} (y_t - \hat{q}_{t}^{(\tau_{m})})
\end{equation*}

\noindent where $ y_t $ is an observation used to forecast evaluation. To evaluate full predictive densities, this score is averaged across all target quantiles for all look ahead time steps using equal weights. A lower QS indicates a better forecast. In some applications it may be needed to have prediction intervals (PIs) and as such, we look at two PI evaluation measures: reliability and sharpness. Reliability, also known as calibration, is a measure which states that over an evaluation set the observed and nominal probabilities should be as close as possible, and the empirical coverage should ideally equal the preassigned probability. Sharpness is a measure of the width of prediction intervals, defined as the difference between the upper $ u_{t}^{\beta_i} $ and lower $ l_{t}^{\beta_i} $ interval values. 

For interval reliability we use the average coverage error (ACE) metric \cite{zhang2014review} and for measuring interval sharpness we use the sharpness score (SS). For measuring reliability, PIs show where future observations are expected to lie, with an assigned probability termed as the PI nominal confidence (PINC) $ 100(1-\beta_i)\% $. Here $ i = 1...M/2 $ indicates a specific coverage level. The coverage probability of estimated PIs is expected to eventually reach a nominal level of confidence over the test data. A measure of reliability which shows target coverage of the PIs is the PI coverage probability (PICP), which is defined by
\begin{equation*}
PICP_i=\frac{1}{N}\sum_{t=1}^{N} \mathbbm{1}\{y_t \in I_{t}^{\beta_i}({x}_t)\}.
\end{equation*}

For reliable PIs, the examined PICP should be close to its corresponding PINC. A related and easier to visualize assessment index is the average coverage error (ACE), which is defined by
\begin{equation*}
ACE = \sum_{\ i = 1}^{M/2} |PICP_i-100(1-\beta_i)|.
\end{equation*}

\noindent This assumes calculation across all test data and coverage levels. To ensure PIs have high reliability, the ACE should be as close to zero as possible. A high reliability can be easily achieved by increasing or decreasing the distance between lower and upper interval bounds. Thus, the width of a PI can also influence its quality. For measuring the effective width of PIs we use the sharpness score proposed by \cite{pinson2007non} which measures how wide PIs are by focusing on the mean size of the intervals only. We define $\hat{q}^{u}_{t}-\hat{q}^{l}_{t}$ as the size of the central interval forecast with nominal coverage rate $ (1 - \beta) $. For lead times $t = 1...N_{test}$, a measure of sharpness for PIs is then given by the mean size of the intervals
\begin{equation*} \label{e:sharp}
SS=\frac{1}{N_{test}}\sum_{t=1}^{N_{test}}(\hat{q}^{u}_{t}-\hat{q}^{l}_{t}).
\end{equation*}

\noindent A lower sharpness score is considered more ideal, but too small and the PIs would not cover enough of the observed data. Thus sharpness is typically a measure to be considered along with reliability and a skill score.


\section{Proposed Methodology} \label{s:mod}

Fourier analysis examines the approximation of functions through their decomposition as a sum or product of trigonometric functions, while Fourier synthesis focuses on the reconstruction of a signal from its decomposed oscillatory components. These well-known processes have a large utility in time series analysis. By decomposing a time series into its frequencies one could then interpolate missing time values by reconstructing the original signal. Further applications include modeling seasonality and even prediction of a time series through extrapolation of an approximated signal. In the application of Fourier analysis for time series analysis, an important method is the discrete Fourier transform (DFT), which converts a series into its frequency domain representation, and the inverse discrete Fourier transform (iDFT) which maps the frequency representation back to the time domain. The transforms can be expressed as either a summation of complex exponentials or sines and cosines by Euler’s formula. In this section we explore existing works on Fourier networks that directly use iDFT in their operation or mimic it, then we describe our proposed FNN methodology for quantile forecasting.

\subsection{Fourier Neural Networks}

Neural networks with sine as an activation function are difficult to train in theory and when initialized randomly yield poor results \cite{parascandolo2016taming}. Thus, few works have attempted to explore Fourier analysis with sinusoidal neural networks. We highlight most of the works here. One of the first FNNs was introduced by Adrian Silvescu \cite{silvescu1997new, silvescu1999fourier} who developed Fourier-like neurons for learning boolean functions. The FNN model used the units of the network to approximate a DFT in its output. Similar in spirit to an FNN was a Fourier transform neural network introduced in \cite{minami1999real} that uses the Fourier transform of the data as input to an artificial neural network. FNNs have since been used for stock prediction \cite{mingo2004fourier}, aircraft engine fault diagnostics \cite{tan2006fourier}, harmonic analysis \cite{germec2009fourier}, and extensions include a single input multiple outputs based FNNs that can turn nonlinear optimization problems into linear ones \cite{mingo2004fourier}, FNNs for output feedback learning control schemes \cite{zuo2009fourier}, and deep FNNs for lane departure prediction \cite{tan2017use}.

There are two recent works that study FNNs for time series prediction that use the Fourier transform of the data as weights. The first is an FNN presented by Gashler and Ashmore in \cite{gashler2014training}. Their technique uses the fast Fourier transform (FFT) to approximate the DFT and then uses the obtained values to initialize the weights of the neural network. Their model uses a combination of sinusoid, linear, and softplus activation units for modeling periodic and non-periodic components of a time series. However, their trained models were slightly out of phase with their validation data. The second study on FNNs for time series prediction is presented by Godfrey and Gashler \cite{godfrey2017neural} who proposed a similar model to \cite{gashler2014training} called neural decomposition (ND), except that they do not use the Fourier transform to directly initialize any weights.  

The ND model is inspired by the inverse discreet Fourier transform where given time $ t $ as input it attempts to model the signal $ x(t) $. However, there are some distinctions between ND and iDFT. First ND allows sinusoid frequencies to be trained and second ND can also model non-periodic components in a signal such as trend. With the ability to train the frequencies, ND learns the actual period of a signal whereas iDFT assumes that the underlying function always has a period equal to the size of the samples it represents. ND is a feedforward neural network with a single hidden layer with $ N $ nodes and has one input and one output node. Hidden nodes are composed of sinusoid units for capturing the periodic component in an underlying signal and other activation functions, such as linear or sigmoid units, for capturing the non-period component. Parameters of ND are initialized in such a way so as to mimic the iDFT. ND is then trained with stochastic gradient descent with backpropagation. ND was applied to time series deterministic forecasting and showed very promising results across different data sets, often beating state-of-the-art methods such as LSTM, SVR, and SARIMA.

\subsection{Quantile Fourier Neural Network}

Inspired by the ND model we propose a new forecasting method which we call the quantile Fourier neural network (QFNN).  Unlike ND and other FNNs our QFNN  model is trained to extrapolate quantiles of an underlying time series, rather than point estimates. We apply sinusoid activation functions to allow QFNN to fit periodic data, and coupled with an augmentation function it is able to probabilistically forecast a time series that is made up of non-periodic components too. The model is defined as follows. Let each $ a_{jk} $ represent an amplitude, each $ \omega_k $ represents a frequency, each $ \phi_k $ represents a phase shift, and  $ b^{[2]}_{\tau} $ and $ b^{[1]}_{k} $ represent bias terms for the quantile signal representation. Let $ f(t) $ be an augmentation function that represents the non-periodic components of the signal. QFNN then can be defined by
\begin{equation}\label{Eq.nd}
q_{t}^{\tau} = f(t) + b^{[2]}_{\tau} + \sum_{k=1}^{N} \left(  a_{\tau, k} \cdot \cos \left( \omega_k t + \phi_k \right) + b^{[1]}_{k}\right) 
\end{equation}

\noindent where given time $ t $ as the input, it attempts to predict the $ \tau $-level quantile. QFNN is loosely modeling a time series as a partial Fourier cosine series
\begin{equation}\label{Eq.FS}
x(t) = A_0 + \sum_{n=1}^{N} A_n \cos(n \omega_0 t + \phi_n )
\end{equation}

\noindent where $ \omega_0 = \frac{2 \pi}{T} $, $ T $ is the period of the signal $ x(t) $, and $ A_0 $, $ A_n $, and $ \phi_n $ are real numbers. The main difference between Eq. \ref{Eq.nd} and Eq. \ref{Eq.FS} is that QFNN does not fix the period of the signal to a predetermined size $ T $, it allows for bias terms, it has an augmentation function to represent non-periodic components of the signal, and it learns the frequencies $ \omega_k $ versus keeping them at a fixed size. The bias terms in the output layer of the network are important because it allows to shift the level of each quantile appropriately.

The hidden layer of QFNN is composed of $ N $ units with a sinusoid activation function and an arbitrary number of units with other activation functions to calculate $ f(t) $. The output layer is composed of $ M $ number of linear units that represent quantiles. The parameters $ a_{jk} $, being the weights between the hidden and output layers allows us to model different amplitudes for composite quantiles while simultaneously learning the frequency and phases for all quantiles in the hidden layer. Utilizing conventional neural network notation $ W^{[1]}  $ is a matrix of the $ f(t) $ unit parameters and the frequency parameters in Eq. 3. The $  b^{[1]} $ vector represents the phases of the sinusoidal components, $ W^{[2]} $ is a parameter matrix of the amplitudes, and we also add additional bias terms to the output nodes for each quantile with the $ b^{[2]} $ vector.

\begin{figure}[h]
	\centering
	\includegraphics[width=1 \textwidth]{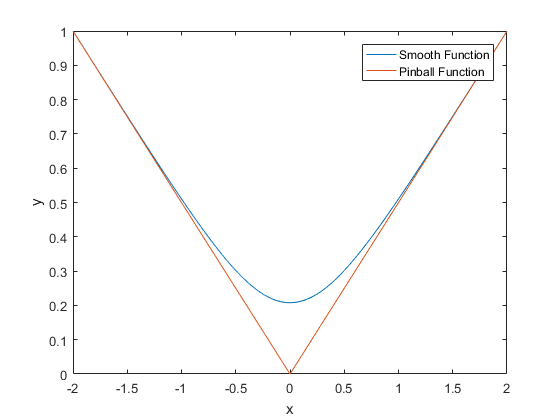}
	\caption{Pinball ball function versus the smooth pinball neural network with smoothing parameter $ \alpha =0.2 $.}
	\label{pinvsmooth}
\end{figure}

\begin{figure}[h]
	\centering
	\includegraphics[width=0.7 \textwidth]{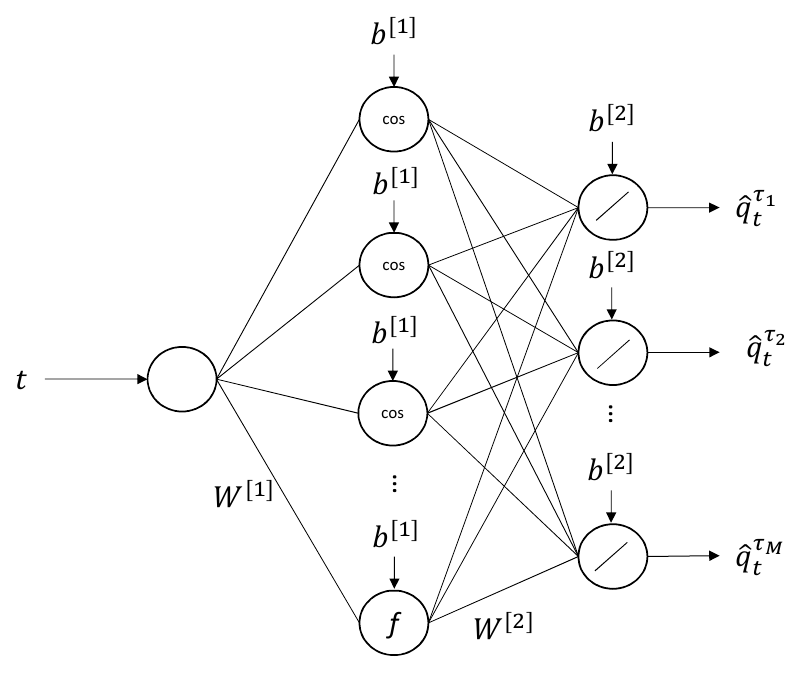}
	\caption{Architecture of the quantile Fourier neural network.}
	\label{f:net}
\end{figure}

To estimate quantiles we need to solve the minimization problem described in Eq. \ref{eq3}. However, the pinball function $ \rho $  in Eq. \ref{eq3}. is not differentiable at the origin, $ x = 0 $. The non-differentiability of $ \rho $ makes it difficult to apply gradient-based optimization methods in fitting the quantile regression model. Gradient-based methods are preferable for training neural networks since they are time efficient, easy to implement and can yield a local optimum. Therefore, we need a smooth approximation of the pinball function that allows for the direct application of gradient-based optimization. A smooth approximation of the pinball function in Eq. \eqref{pinball}  is proposed by Zheng in \cite{zheng2011gradient} as
\begin{equation}
S_{\tau,\alpha}(u) = \tau u + \alpha \log \left( 1 + \exp \left(-\frac{u}{\alpha}\right) \right), 
\end{equation}

\noindent where  $ \alpha>0 $ is a smoothing parameter and $ \tau \in [0,1] $ is the quantile level we're attempting to estimate. In Fig. \ref{pinvsmooth} we see the pinball function with $ \tau = 0.5 $ as the red line and the a smooth approximation as the blue line with $ \alpha = 0.2 $. Zheng proves \cite{zheng2011gradient} that in the limit as $ \alpha \rightarrow 0^{+} $ then $ S_{\tau,\alpha}(u) = \rho_{\tau}(u) $. With this smooth approximation we can then define the cost minimization problem for QFNN as
\begin{equation} \label{opti}
\begin{multlined}
E = \frac{1}{NM}\sum_{t=1}^{N} \sum_{m = 1}^{M}
\left[ \tau_m (y_t - \hat{q}_{t}^{(\tau_m)}) + \alpha \log \left( 1 + \exp \left( - \frac{y_t-\hat{q}_{t}^{(\tau_m)}}{\alpha} \right) \right) \right]. 
\end{multlined}
\end{equation}
where $ M $ number of $ \tau $'s we are trying to estimate in the output layer. The input to hidden neurons is calculated, in vectorization notation, by $ Z_{t}^{[1]} =  W^{[1]} t + b^{[1]} $, the output of the hidden layer then uses the logistic activation function $ H_{t} = \cos \left(  Z_{t}^{[1]} \right), f\left(  Z_{t}^{[1]} \right) $. The input to output neurons is then calculated by $ Z_{t}^{[2]} =  W^{[2]} H_t + b^{[2]} $, and the output layer uses the identity activation function $ \hat{Q}_t =  Z_{t}^{[2]} $ where $ \hat{Q}_t $ is a vector output of the estimated composite quantiles. An architectural view of the QFNN is shown in Fig. \ref{f:net}.

\subsection{Implementation Details}

The proposed QFNN model is trained using gradient descent with backpropagation. The training process allows the model to learn better frequencies and phase shifts so that the sinusoid units more accurately represent the seasonality of an underlying time series. Since frequencies and phase shifts can change, the model learns a more reliable periodicity of the underlying series rather than assuming the period is of a predetermined size. Training also tunes the weights of the augmentation function which estimates the non-periodic component of the time series. 

In addressing the problem of overfitting, the ND model proposes the use of L1 regularization on the output weights to promote sparsity and shrink the less essential sinusoidal components. However, ND is designed for estimating a single output - the expected value of the time series. In our QFNN model we are estimating multiple quantile outputs where each cosine component may play a different role in the estimation of each quantile. Thus we do not want to shrink cosine components, but are instead interested in learning different possible models during the training phase to see how each cosine component affects individual estimates. To achieve such regularization we use dropout \cite{srivastava2014dropout}. 

Dropout is a recently proposed regularization technique that randomly drops units, along with their connections, from the neural network during training. This can significantly reduce overfitting and gives major improvements over other regularization methods. Our main proposal for the usage of dropout in QFNN vs L1 regularization is to efficiently approximate in training all possible sub-models of a given architecture and then take their average. No regularization was applied for the augmentation function $ f(t) $.

Next, for parameter initialization, there is a considerable distinction in how QFNN is initialized compared to other FNNs. Instead of initializing parameters to mimic the iDFT we randomly set all bias terms near 0, the output weights $ W^{[2]} $ which represent the amplitudes are initialized randomly close to 1, and the input weights $W^{[1]}$ which represent the frequencies are set to multiples of $ \pi k $ where $ k $ is a specific hidden node. The input weight parameters of the augmentation function $ f(t) $ are initialized randomly to close to 1 and its bias terms randomly set near 0. 

Before training starts, the input data is preprocessed in the same fashion as in \cite{godfrey2017neural} to improve learning. First, the time associated with each training sample is normalized between 0 (inclusive) and 1 (exclusive) on the time axis. By doing this normalization testing data points will have a time value greater than or equal to 1. With this normalization the 1/N term in the frequencies is taken into account by transforming t into t/N. Next, if the max value in the training set is greater then 10, then the training set is scaled between 0 and 10. This normalization avoids any look ahead bias by using on the max value in the training set only. Both these preprocessing steps expedite learning and help prevent the model from falling into local optimums.

\begin{figure}[h]
	\centering
	\includegraphics[scale=1 ]{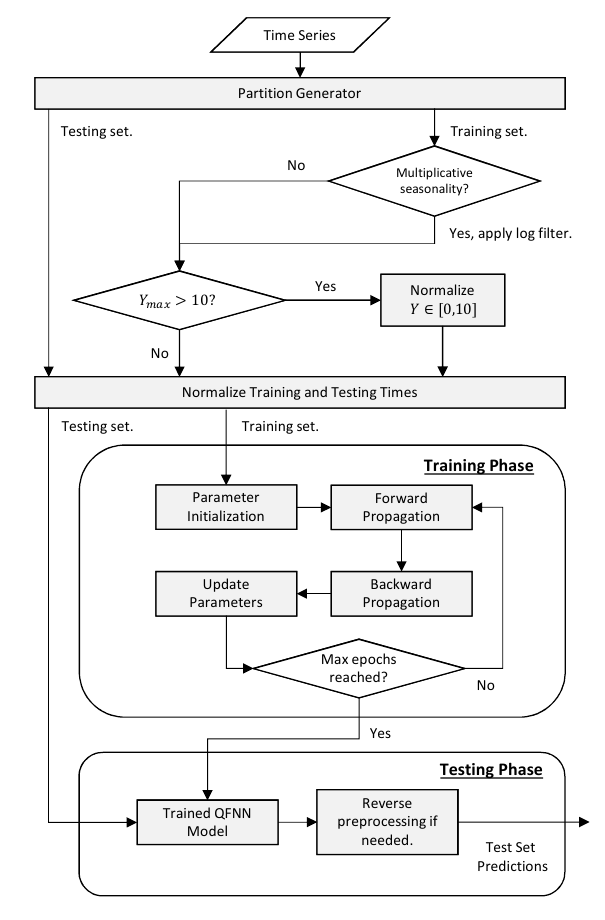}
	\caption{Flowchart of proposed methodology using QFNN.}
	\label{f:flowchart}
\end{figure}

We present the full proposed QFNN methodology architecture in Fig. \ref{f:flowchart}. Summarizing our methodology, we first partition a given time series into training and testing sets. Preprocessing of the training set is then conducted which includes applying a logarithmic filter if multiplicative seasonality is present. A multiplicative time series can be identified by visually inspecting a plot of the series or by analyzing it's autocorrelation plot. This is also considered a key step in the Box–Jenkins decomposition method \cite{brockwell2002introduction}. This preprocessing step is needed since QFNN is an additive model. Next if the training data has points above 10, the data is normalized between 0 and 10. Next, the training and testing times are normalized so that training times are between 0 and 1. Parameters of the QFNN are then initialized as described in the previous paragraphs. Training of the model is conducted using batch gradient descent. After the max number of training epochs is reached the QFNN model is ready to be used for testing. Forecasts can be provided for a multi-step period of indefinite time steps. After a test set prediction is made, preprocessing steps are reversed if any were conducted. Preprocessing steps may include scaling a time series back to its original scale or removing the logarithmic filter from the outputs by exponentiating the predictions. We implement QFNN in Python 3.6 using the Keras/tensorflow framework \cite{chollet2015keras}.


\section{Validation} \label{s:res}

In this section, we evaluate the effectiveness of the QFNN method for the estimation of quantiles and prediction intervals on 8 diverse datasets, described in Section \ref{casestudies}, and compare results with 9 benchmark methods described in Section \ref{Benchmark}. For each case (dataset), we conduct three kind of studies to see how the proposed algorithm handles. We analyze the ability to estimate single median percentile values, multiple percentiles (which are used combined to form multiple prediction intervals), and extremely low/high percentiles. For experimentation in each case study, we use 50\% of the data for training and the other 50\% for testing. All models only saw observations in the training sets. Test data were never presented to the models and were used just to calculate evaluation metrics. The use of 50\% of the time series for testing was done to achieve the goal of long-term multi-step PF. 

For the analysis of the estimation of median percentiles, of nominal level $ \tau = 0.5 $, we use QS for evaluation. For the analysis of estimating multiple percentiles, we look at 100 quantiles whose nominal values are equally spaced between 0 and 1 (0.0099, 0.01988, 0.0297, ..., 0.9702, 0.9801, 0.9900). These quantiles can be combined to form 50 prediction intervals with nominal coverage rates of 0.95\% to 98.01\%. We also analyze extreme valued percentiles and prediction intervals whose nominal coverage is very low or high. We estimate ten low and high valued quantiles (with levels of 0.005, 0.010, 0.015, 0.02, 0.025, 0.975, 0.98, 0.985, 0.99, 0.995) which combined together form five prediction intervals with nominal coverage of 95\%, 96\%, 97\%, 98\%, and 99\%. For analyzing multiple and extreme percentile estimates we use QS, and for analyzing prediction interval performance we use ACE and SS.

\begin{figure}[h]
	\centering
	\includegraphics[width=1 \textwidth]{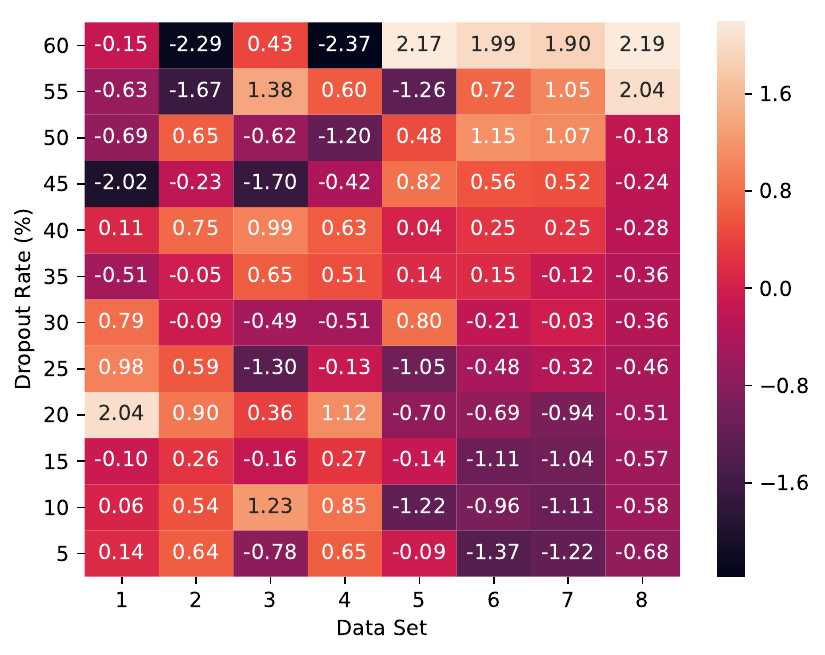}
	\caption{Heat-map of the grid search results for dropout rate in the QFNN model on estimating median values. QS measures have been standardized to show comparison between datasets.}
	\label{grid_search_median}
\end{figure}

In each of the experiments, we use one linear activation function for the augmentation function of QFNN to capture the trend. We found that using more than one augmentation function or using nonlinear activation functions such as tanh, sigmoid, and rectified linear units did not provide any significant improvements in trend estimation. The dropout regularization rate can have a large impact on the performance of QFNN in terms of overfitting. Therefore, we apply grid search using the training data for validation. We pick an optimal value of dropout, which we search in a range between 5\% to 60\%, based on the lowest QS. Grid search is applied for each case study. An example grid search is seen in Fig. \ref{grid_search_median} for choosing the dropout rate for estimating the median percentile for each dataset. The QS metric is standardized to allow for comparison between datasets.

For all experiments, we use a maximum training iteration of 10,000 for QFNN and a smoothing rate of 0.01. The learning rate of 0.5 was used in all experiments, except when estimating the single median value for the Air Passengers dataset which failed to converge during training. In that case, a lower learning rate of 0.1 was used. Through initial analysis of training results, we found that altering the training iterations, smoothing rate, or learning rate had little impact, so no form of hyperparameter tuning was used for them. 

\subsection{Case Studies} \label{casestudies}

We carry out experiments on eight nonstationary univariate time series datasets, seven being real-world case studies and one synthetic case. These datasets were explicitly picked because they display a diverse set of periodic and aperiodic patterns such as trend, additive and multiplicative seasonality, multiple seasonality, cycles, and irregular patterns. Table \ref{t:cases} the characteristics of all the datasets. The first case study is the classical Air Passengers time series \cite{air} which is composed of 144 samples of the number of passengers flying each month from January 1949 to December 1960. It has a positive linear trend and multiplicative seasonality.

The second case study is the yearly mean and monthly smoothed total sunspot numbers from 1700 to 2017 \cite{sidc}. It consists of 318 samples with a time granularity of one year. This time series includes an unstable (non-constant) seasonal patterns over time. Case study 3 is the load demand from ISO New England \cite{load}. Its time series is composed of 744 samples for January 2017. Target values represent real-time demand in MW for wholesale market settlement from revenue quality metering. This case study displays seasonal and cyclical patterns. Internet traffic data in bits from a private ISP with centers in 11 European cities is used for the 4rth case study \cite{internet}, which exhibits multiple seasonality. We use the data that corresponds to a transatlantic link and was collected from on June 18 to July 16, 2005. 

The highly random movements of the stock market are almost impossible to predict but some stocks may exhibit unseen cycles or trends over more extended periods of time \cite{fama1988business}. To examine such possible patterns we use the closing stock prices of Apple Inc. over five years from 2012 to the beginning of April 2018 \cite{aapl}. The next two case studies are normalized solar and wind power for September 2012 and January 2012. These two datasets come from the Global Energy Forecasting Competition of 2014 \cite{hong2016probabilistic}. Wind power is highly chaotic and is very difficult to forecast from univariate time series.

The last case study looks at ocean wave elevation, the main motivation for using such data is the irregular sinusoidal nature of waves. Due to the difficulty of finding high resolution deep ocean wave elevation measurements we construct a synthetic dataset. For simulating of ocean waves we focus on vertical sensors for predicting irregular wave formations. Under generally well accepted assumptions \cite{falnes2002ocean}, the wave elevation for sensor locations $(x,y)$ on the ocean surface for all times $t$ the exact time waveform which would be observed at a particular point in the ocean can be described by
\begin{equation} \label{eq:el}
H (x,y,t)= \sum_{i=1}^{L} A_i \cos\bigg(\frac{\omega_{i}^{2}}{g}\big(x\cos(\beta_i)+y \sin(\beta_i)\big)- \omega_i t + \phi_i \bigg),
\end{equation}

\noindent which has the parameters $A$ for the amplitude, $\omega$ for the frequency measured in radians per second $(rads/s)$, $\beta$ for the wave angular direction in radians measured relative to the x-axis, and $\phi$ for the phase in radians. We chose to estimate waves at the origin with $ L = 2 $, and for each parameter we arbitrarily chose the values $ A = [1,1.5] $, $ \omega = [0.5\pi, 0.1 \pi] $, and $ \phi = [1.2, 1.4] $. To each observation we also include additive white Gaussian noise which we assume come from the sensors.

\subsection{Benchmark Methods} \label{Benchmark}

To thoroughly examine the forecasting accuracy of our QFNN method we compare it to nine simple and state-of-the-art probabilistic forecasting methods. These include two naive approaches, the uniform and persistence methods. Three-time series models which are the autoregressive integrated moving average model, the seasonal autoregressive integrated moving average model, and exponential smoothing with trend and seasonality. Lastly, we use four advanced PF methods: linear quantile regression, polynomial quantile regression, composite support vector quantile regression, and a composite quantile regression neural network. All methods are implemented in Python 3.6.

The uniform method (UM), commonly used in wind power PF studies\cite{sideratos2012probabilistic}, is the simplest of all the methods. UM assumes that any observation in the time series has equal probability to occur at any time step. The support of the UM is defined by the parameters $ a $ and $ b $ which are the minimum and maximum values of the training set for each case study. Quantiles are then defined by $ F^{-1}(\tau) = (1-\tau)a + \tau b $ for $\tau \in [0,1]$. For deterministic forecasting, the persistence forecast method is a very popular benchmark and is known to be hard to outperform for single point or short look-ahead forecasts. We use the persistence method (PM) \cite{wan2014optimal} for PF as a benchmark where the forecast error is assumed to be random and normally distributed, it's mean and variance are computed by the latest observations. For our experiments, we use the last $ S $ observations from the training set to calculate the moments of the PM distribution where $ S $ corresponds to the size of the seasonality derived from the autocorrelation function (ACF). To ensure that UM and PM can estimate appropriate multi-step forecasts we extend both of them by adding an estimated linear trend component from the training data. 

The next three benchmark methods are well established time series models. We use the autoregressive integrated moving average (ARIMA) model, seasonal ARIMA (SARIMA) model, and exponential smoothing with trend and seasonality (ETS) model, also known as the Holt-Winters seasonal method. We choose ARIMA because it can eliminate non-stationarity through an initial differencing step to better fit time series for prediction, and we select SARIMA and ETS to capture periodic patterns better. Parameters of ARIMA and SARIMA are selected using grid search with the Akaike information criterion and the application of the parsimony principle to prevent over-fitting. The seasonal parameter $ S $ for SARIMA and ETS is chosen using the ACF. Quantiles are estimated for ARIMA, SARIMA, and ETS assuming the normality assumption \cite{arima,hyndman2002state}. ARIMA and SARIMA are implemented in Python using the sarimax function from the statsmodels package \cite{sarimax}, and ETS is implemented in Python from a holtwinters package \cite{holt}. 

Assuming a normal distribution for quantile prediction with ARIMA, SARIMA, and ETS is a parametric PF approach and can be somewhat restrictive and may not appropriately estimate the forecast distribution. Therefore, we use four advanced nonparametric PF methods linear quantile regression (QR), polynomial QR (PQR), composite support vector quantile regression (SVQR) \cite{hatalis2018empirical}, and composite quantile regression neural network (QRNN) \cite{hatalis2017smooth}. All four of these advanced benchmarks use time as their input for direct comparison to QFNN. QR, PQR, and QRNN are implemented in Keras/tensorflow \cite{chollet2015keras}. We implement composite SVQR using the liquidSVM package with quantile regression loss and Gaussian kernel \cite{steinwart2017liquidsvm}. Our last benchmark, QRNN, is similar in structure to our QFNN model where we use one hidden layer, rectified linear units in the hidden layer, and uses a smooth approximation of the pinball loss function. The main differences in QRNN are that L2 regularization is used, no augmentation function is used, and all parameters are randomized during initialization. For QRNN we use a maximum training iteration of 10,000, a learning rate of 0.1, and a smoothing rate of 0.01, and a L2 regularization rate of 0.001 for all experiments.

\subsection{Results and Discussion}

\subsubsection{Median Percentile Analysis}

\begin{figure}[h]
	\begin{center}
		\includegraphics[width=1\textwidth]{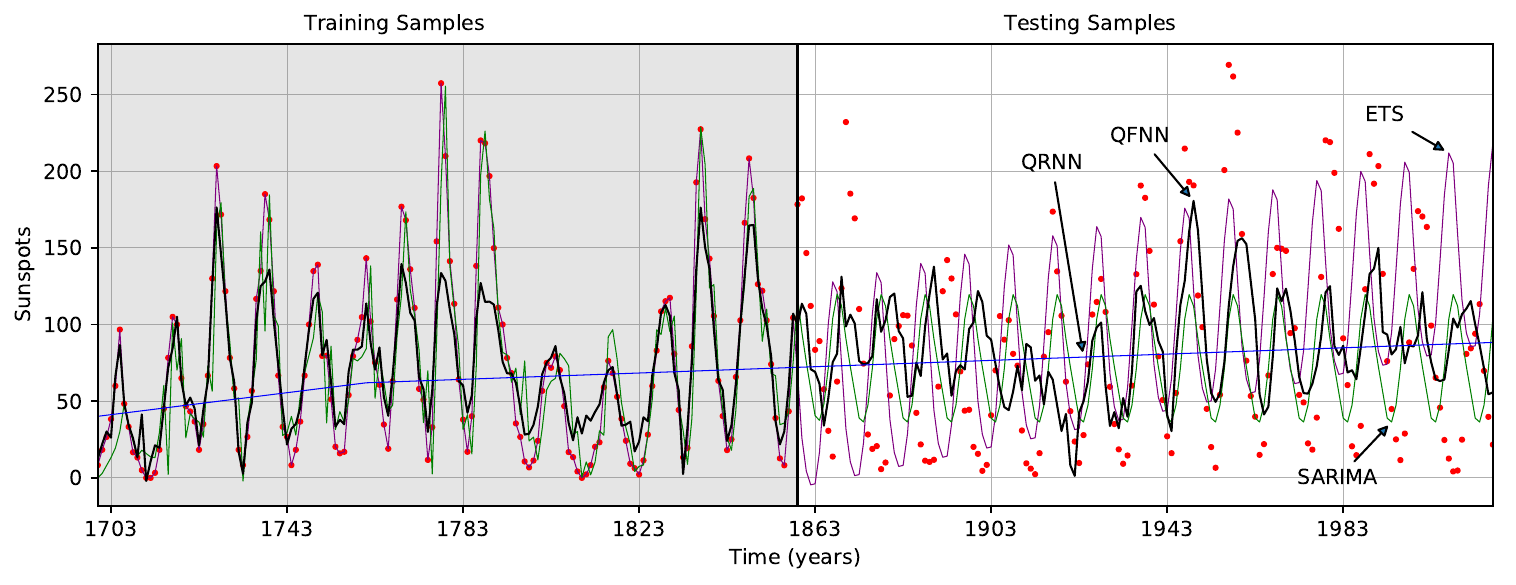}
	\end{center}
	\caption{Forecast comparison of the median quantile for the Sunspots time series (red dots) by QFNN (shown in black), QRNN (shown in blue), SARIMA (shown in green), and ETS (shown in purple). QRNN fails to capture any meaningful pattern in its prediction. SARIMA captures a seasonal pattern that is out of phase with the sunspot test series, and ETS shots off in the test set with some seasonal pattern and positive trend. QFNN captured a periodic pattern that is a bit more in phase with the number of sunspots over the years and is also able to learn multiple seasons of sunspots thus providing the most accurate quantile forecast of all the methods.}
	\label{sun_median}
\end{figure}

\begin{figure}[h]
	\begin{center}
		\includegraphics[width=1\textwidth]{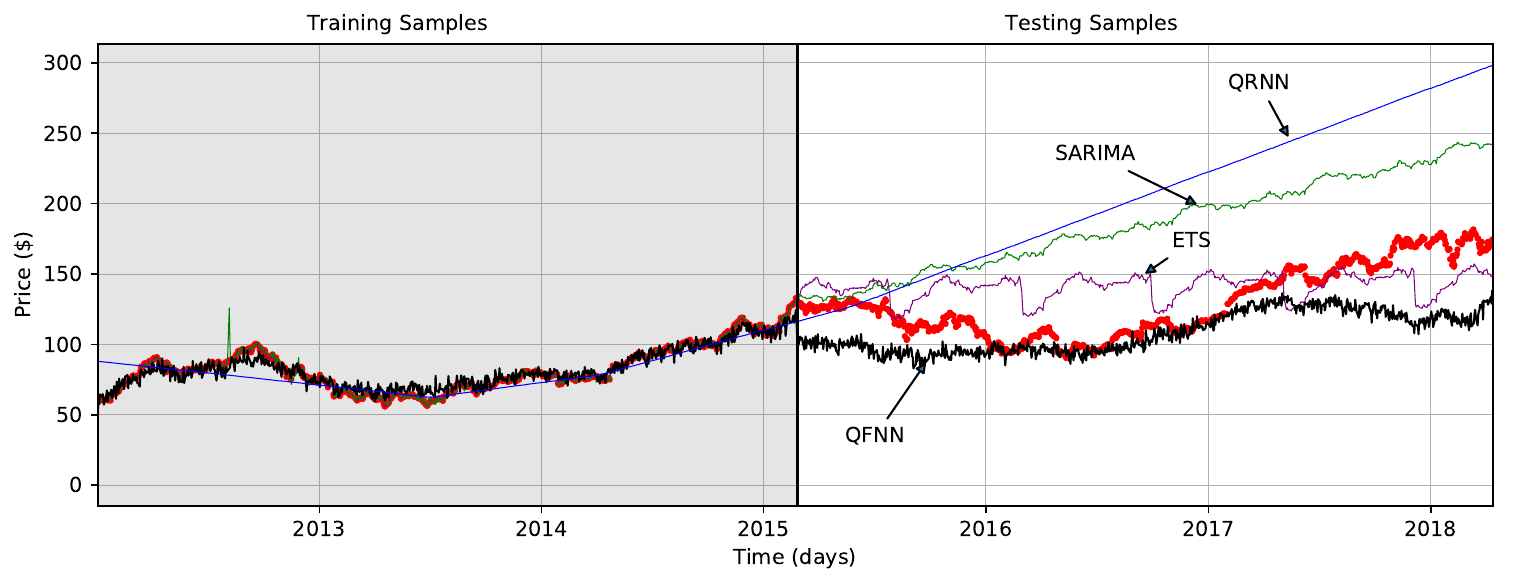}
	\end{center}
	\caption{Forecast comparison of the median quantile for the Apple Closing Stock Price time series (red dots) by QFNN (shown in black), QRNN (shown in blue), SARIMA (shown in green), and ETS (shown in purple). QRNN learns the linear trend of the stock series but nothing else. ETS learns a non-existing seasonal pattern, and SARIMA does not seem to capture any pattern beyond trend. We can see in the plot that QFNN learns a cyclic trend of the stock price which follows the test set better than any other method.}
	\label{stock_median}
\end{figure}

\begin{figure}[h]
	\begin{center}
		\includegraphics[width=1\textwidth]{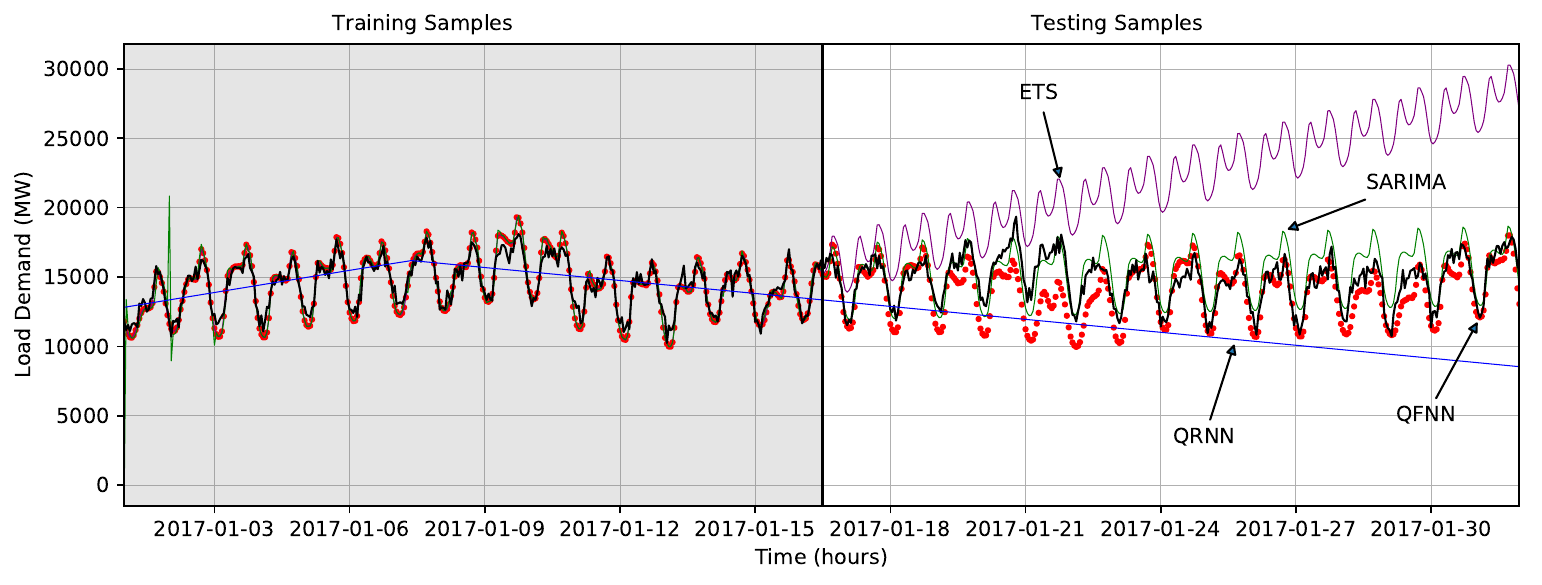}
	\end{center}
	\caption{Forecast comparison of the median quantile for the Load Demand time series (red dots) by QFNN (shown in black), QRNN (shown in blue), SARIMA (shown in green), and ETS (shown in purple). QRNN captures a poor and small cyclic pattern in the training but fails to repeat that cyclic pattern in the testing data. SARIMA captures the seasonality but is unable to capture any cycles in the test set, and ETS shots off in the test set with a positive trend. QFNN  learns both the seasonal and cyclical pattern of the load demand.}
	\label{load_median}
\end{figure}

\begin{figure}[h]
	\begin{center}
		\includegraphics[width=1\textwidth]{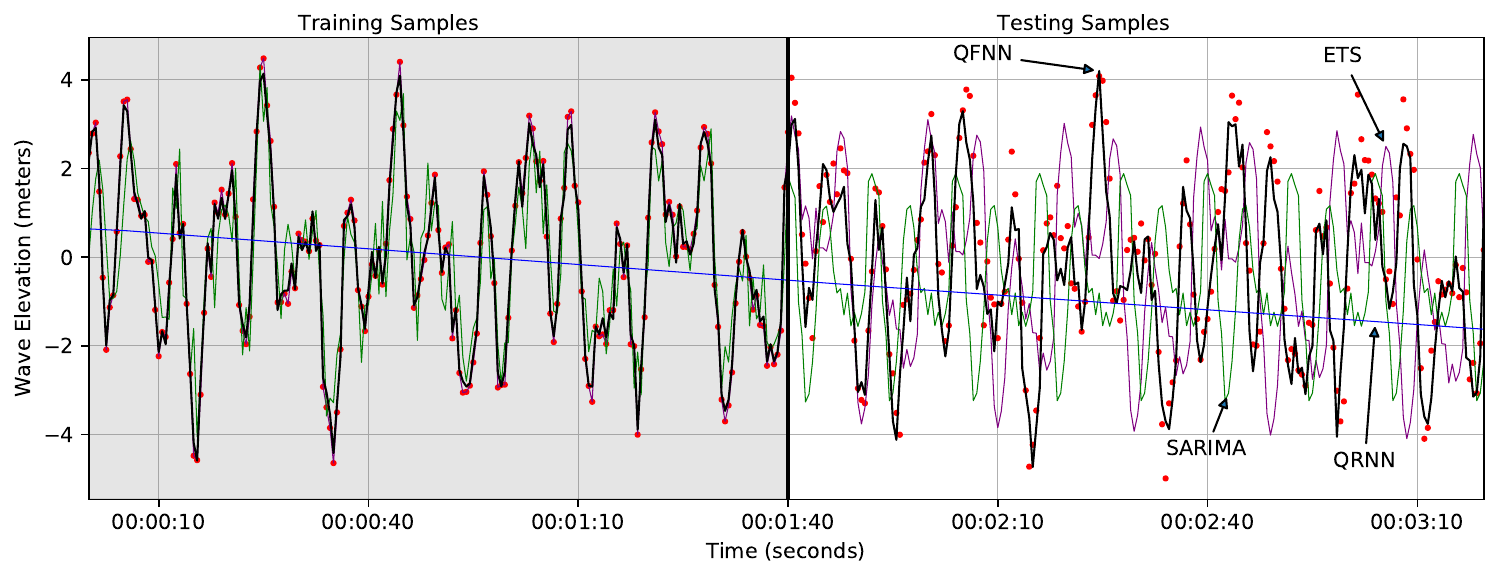}
	\end{center}
	\caption{Forecast comparison of the median quantile for the Wave Elevation time series (red dots) by QFNN (shown in black), QRNN (shown in blue), SARIMA (shown in green), and ETS (shown in purple). QRNN does not fit any pattern in the training or testing samples. SARIMA fits a seasonal pattern but is unable to properly estimate phase or amplitude of the periodicity of wave elevation. ETS estimates a poor similar seasonal fit as SARIMA. QFNN estimates a near perfect fit in both the training and testing samples.}
	\label{solar_median}
\end{figure}

\begin{figure}[h]
	\begin{center}
		\includegraphics[width=1\textwidth]{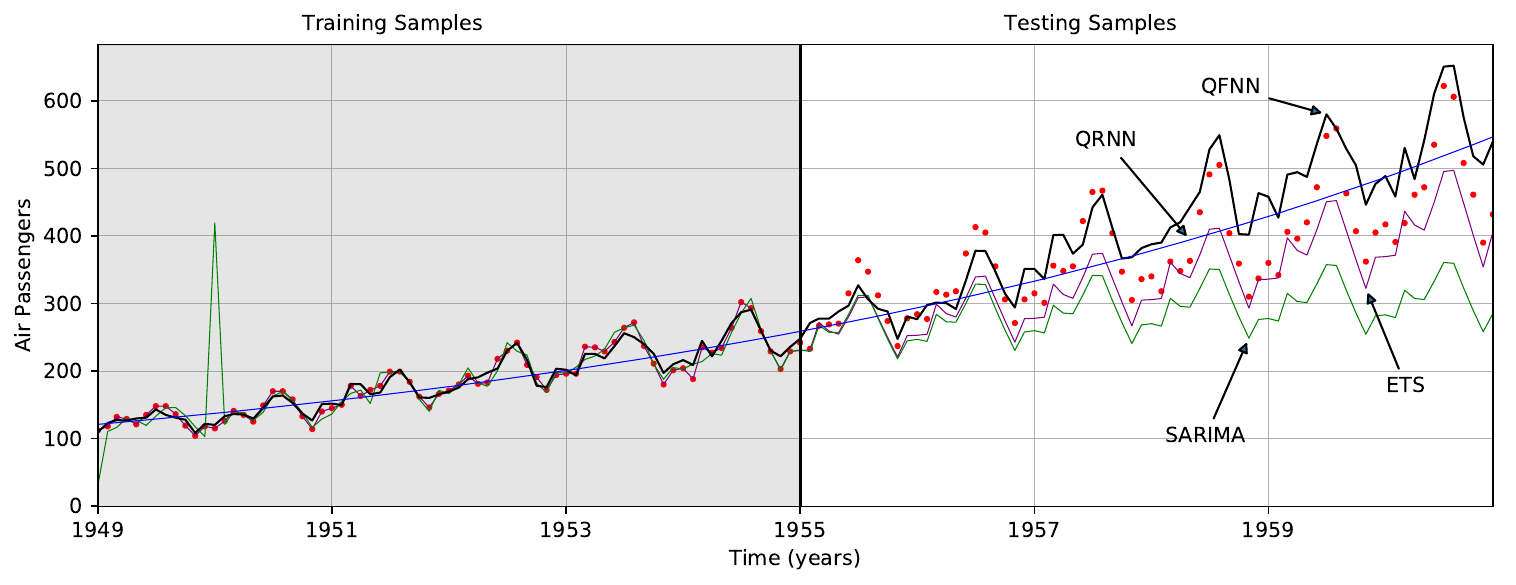}
	\end{center}
	\caption{Forecast comparison of the median quantile for the Air Passengers time series (red dots) by QFNN (shown in black), QRNN (shown in blue), SARIMA (shown in green), and ETS (shown in purple). SVQR estimates the trend but not the seasonality so well. ETS and SARIMA estimate both trend and seasonality well, but the median forecasts fall below and above the test data. QFNN learns the shape of the data better and appropriately captures the median.}
	\label{air_median}
\end{figure}

\begin{figure}[h]
	\centering
	\includegraphics[width=1 \textwidth]{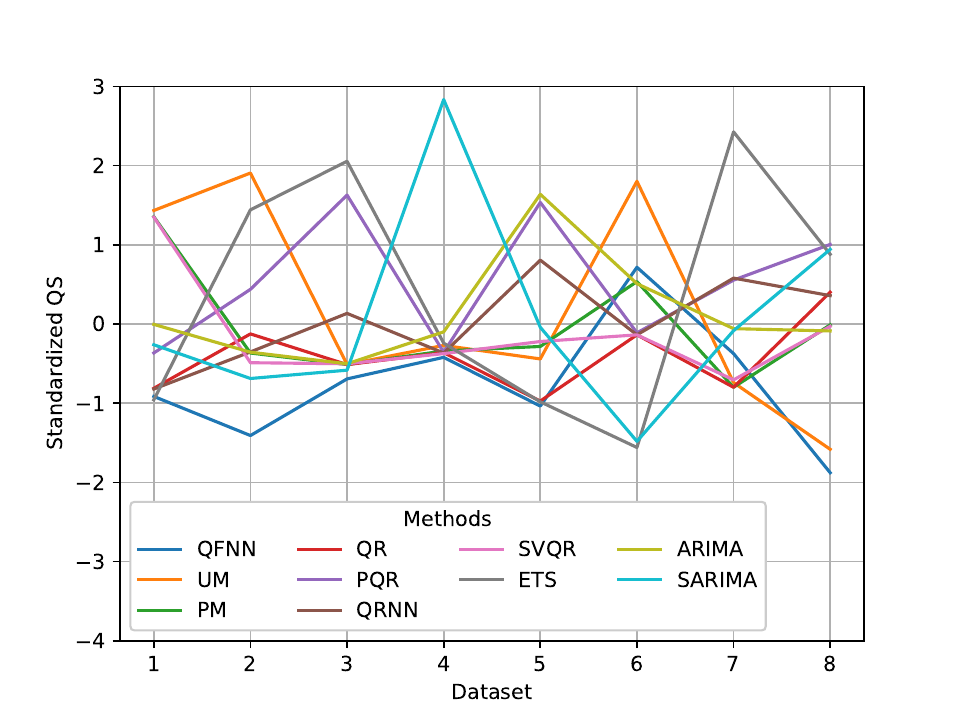}
	\caption{Standardized QS for QFNN and benchmark methods across the 8 datasets for estimating median percentile. QFNN yields the lowest scores for each dataset except for solar and wind power where ETS and QR had the lowest scores respectively.}
	\label{median_QS}
\end{figure}

We first analyze the performance of QFNN and benchmark methods for predicting the median percentile for each case study. We utilize QS for evaluation between benchmarks (on the test component of each dataset), and standardized QS to see result comparisons across case studies. From Fig. \ref{median_QS} we see the standardized QS for every method. QFNN tied closely to ETS with a nonstandard QS of 22.25 for QFNN and 20.82 for ETS. QFNN resulted in the lowest QS for every dataset except for solar and wind power where PM and ETS had better performance. For figures \ref{sun_median} to \ref{air_median} we plot the estimated median percentiles where red dots represent the underlying case study time series. The colored curves represent median forecasts of QFNN and 3 benchmark methods that were able to capture the most information of periodic and aperiodic patterns visually. We found that only the benchmark methods SARIMA and ETS for case studies 1 to 3 and 7 to 8 in Table \ref{t:cases} were able to fit meaningful periodic patterns and thus we only display these three benchmarks. We also include QRNN, a quantile regression based benchmark for comparison to QFNN with the same loss function. Most benchmark methods yielded poor fits for case studies 4, 6, and 7; therefore we do not show these plots. We discuss a visual evaluation of individual case studies in Section \ref{s:individual}.

\subsubsection{Multiple and Extreme Percentile Estimation}

\begin{figure}[h]
	\begin{center}
		\includegraphics[width=1\textwidth]{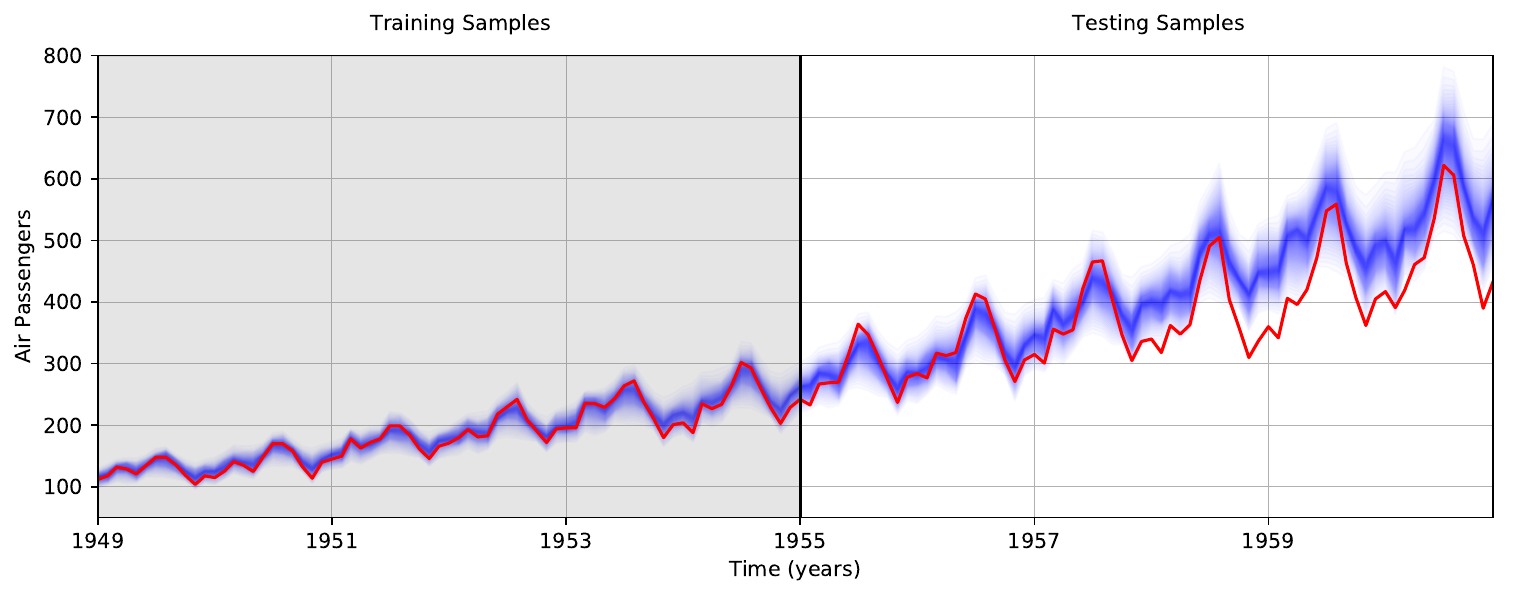}
	\end{center}
	\caption{Probabilistic forecasting of 50 prediction intervals for the Air Passengers series.}
	\label{air_PIs}
\end{figure}

\begin{figure}[h]
	\begin{center}
		\includegraphics[width=1\textwidth]{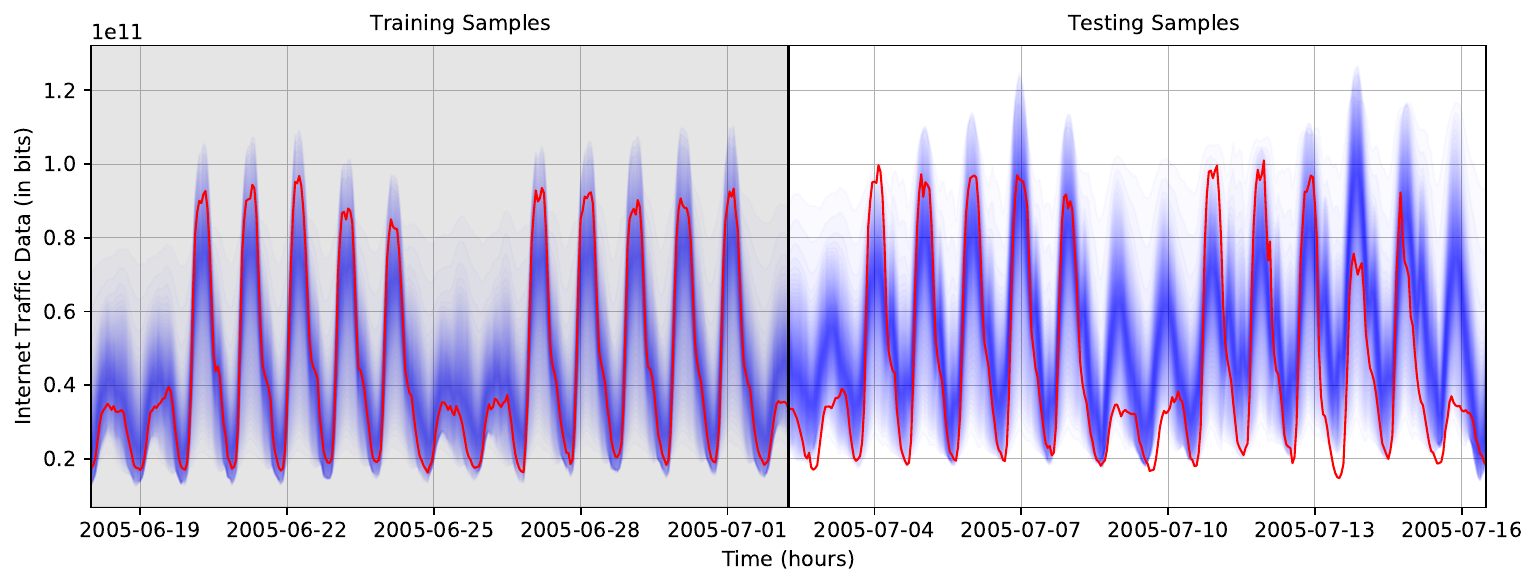}
	\end{center}
	\caption{Probabilistic forecasting of 50 prediction intervals for the Internet Traffic series.}
	\label{internet_PIs}
\end{figure}

\begin{figure}[h]
	\begin{center}
		\includegraphics[width=1\textwidth]{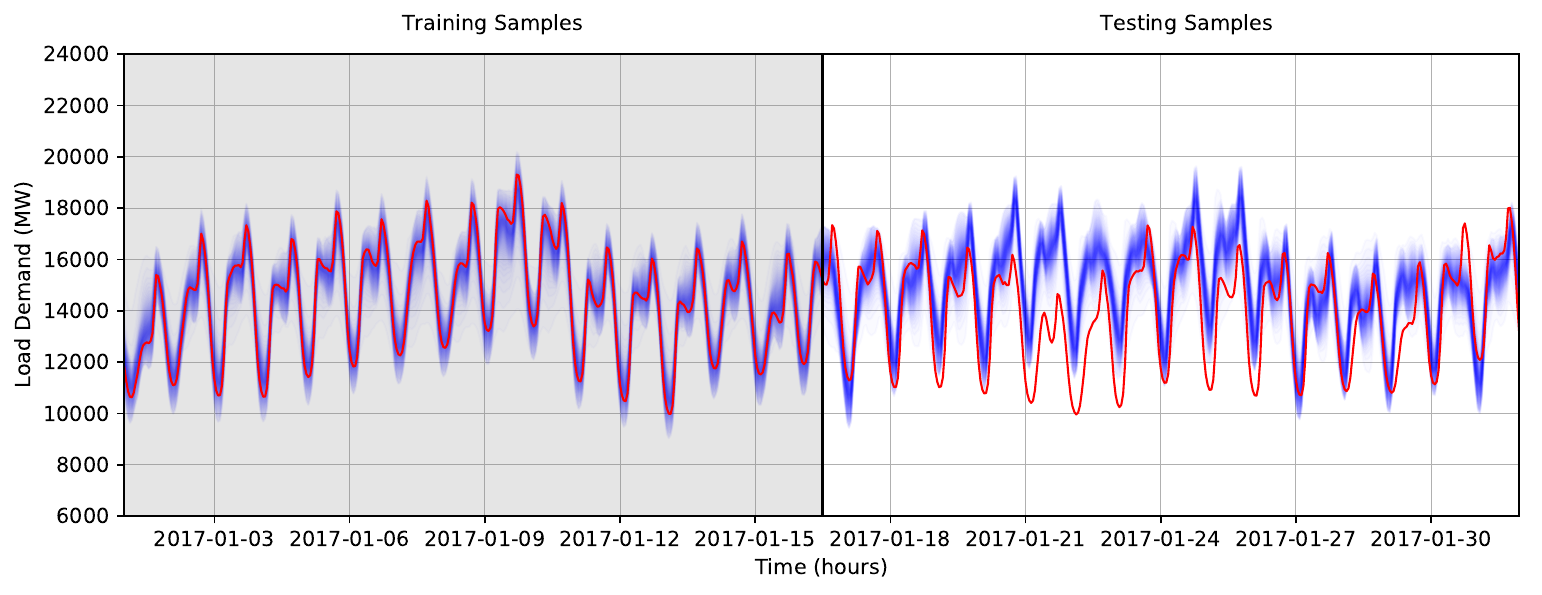}
	\end{center}
	\caption{Probabilistic forecasting of 50 prediction intervals for the Load Demand series.}
	\label{load_PIs}
\end{figure}

\begin{figure}[h]
	\begin{center}
		\includegraphics[width=1\textwidth]{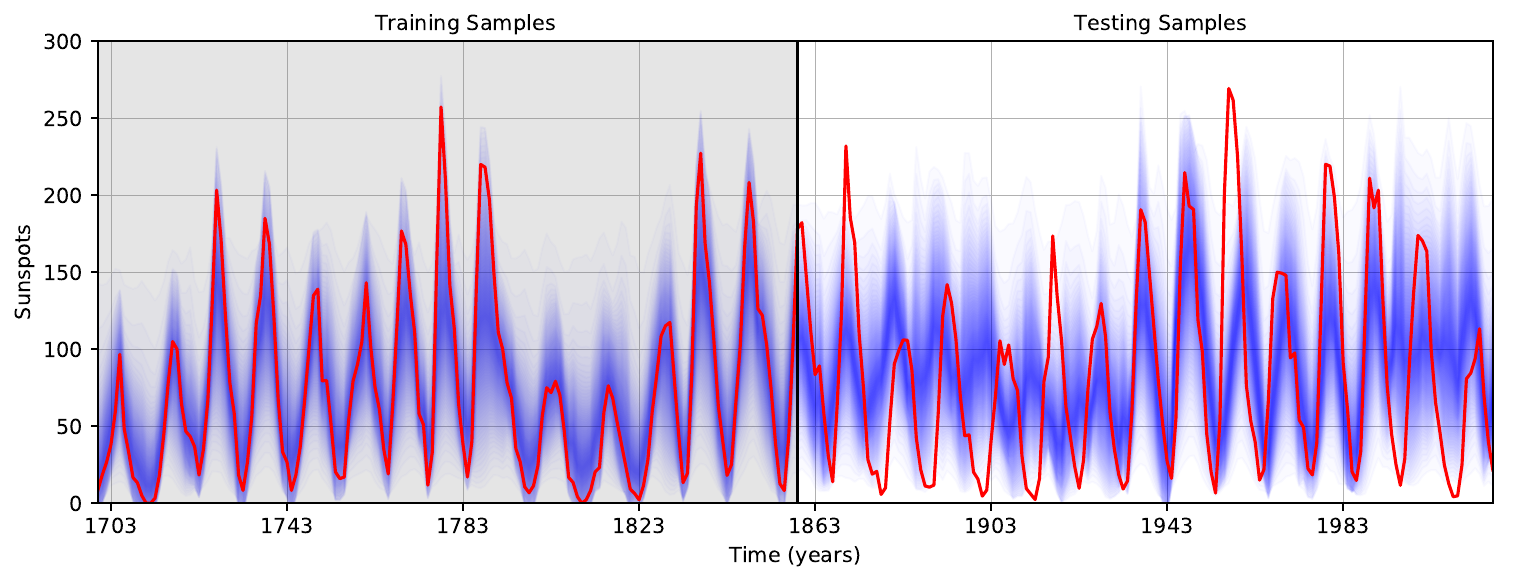}
	\end{center}
	\caption{Probabilistic forecasting of 50 prediction intervals for the Solar Power series.}
	\label{solar_PIs}
\end{figure}

\begin{figure}[h]
	\begin{center}
		\includegraphics[width=1\textwidth]{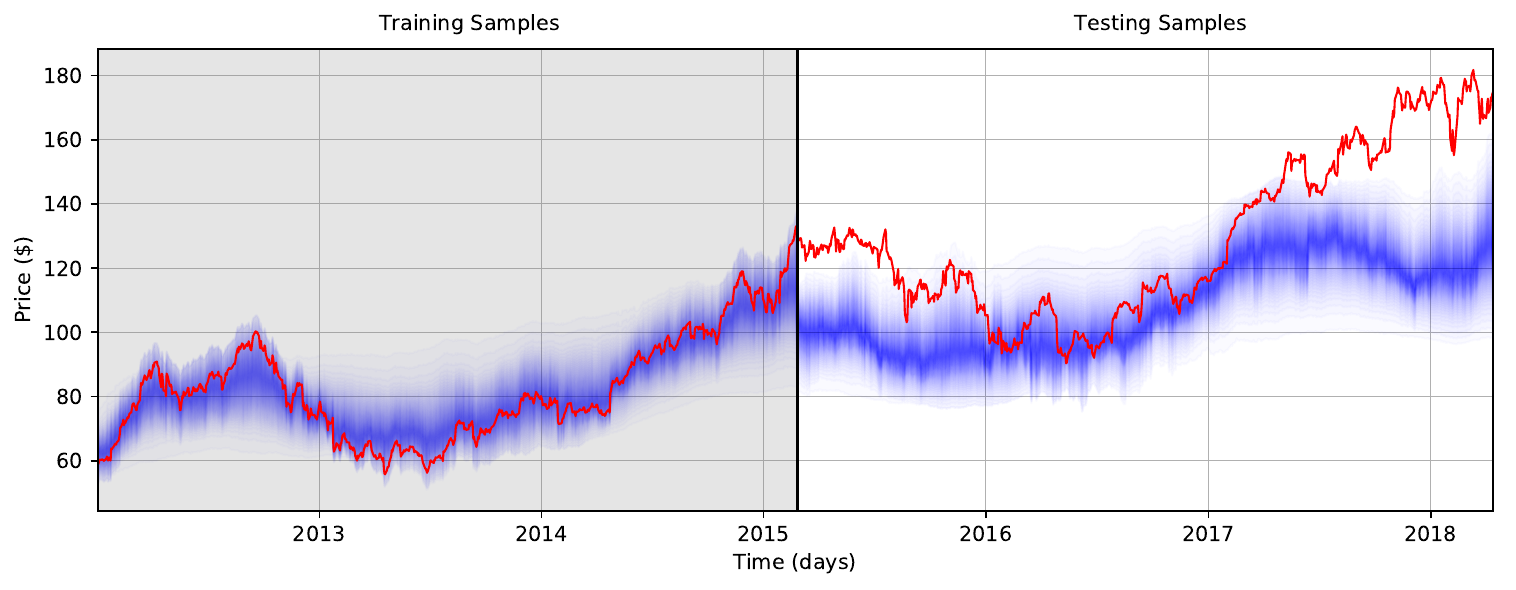}
	\end{center}
	\caption{Probabilistic forecasting of 50 prediction intervals for the Apple Closing Stock Prices time series.}
	\label{stock_PIs}
\end{figure}

\begin{figure}[h]
	\begin{center}
		\includegraphics[width=1\textwidth]{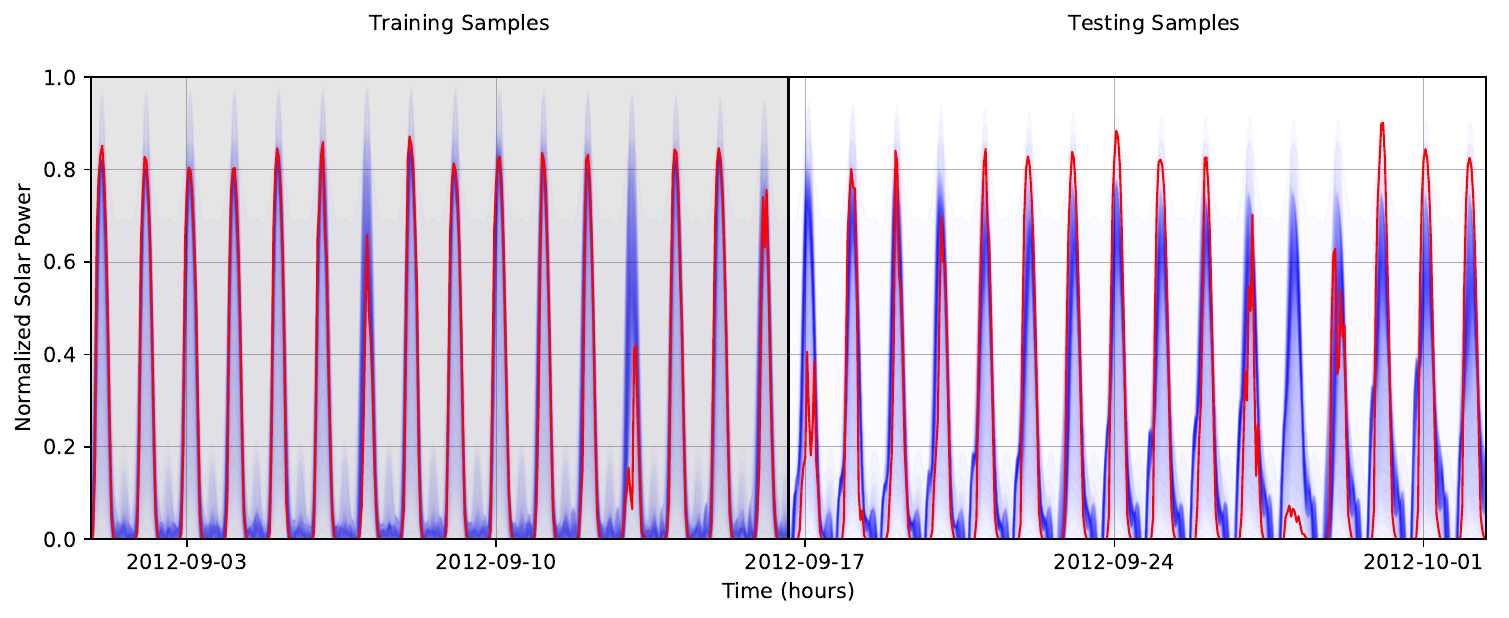}
	\end{center}
	\caption{Probabilistic forecasting of 50 prediction intervals for the Sunspots time series.}
	\label{sun_PIs}
\end{figure}

\begin{figure}[h]
	\begin{center}
		\includegraphics[width=1\textwidth]{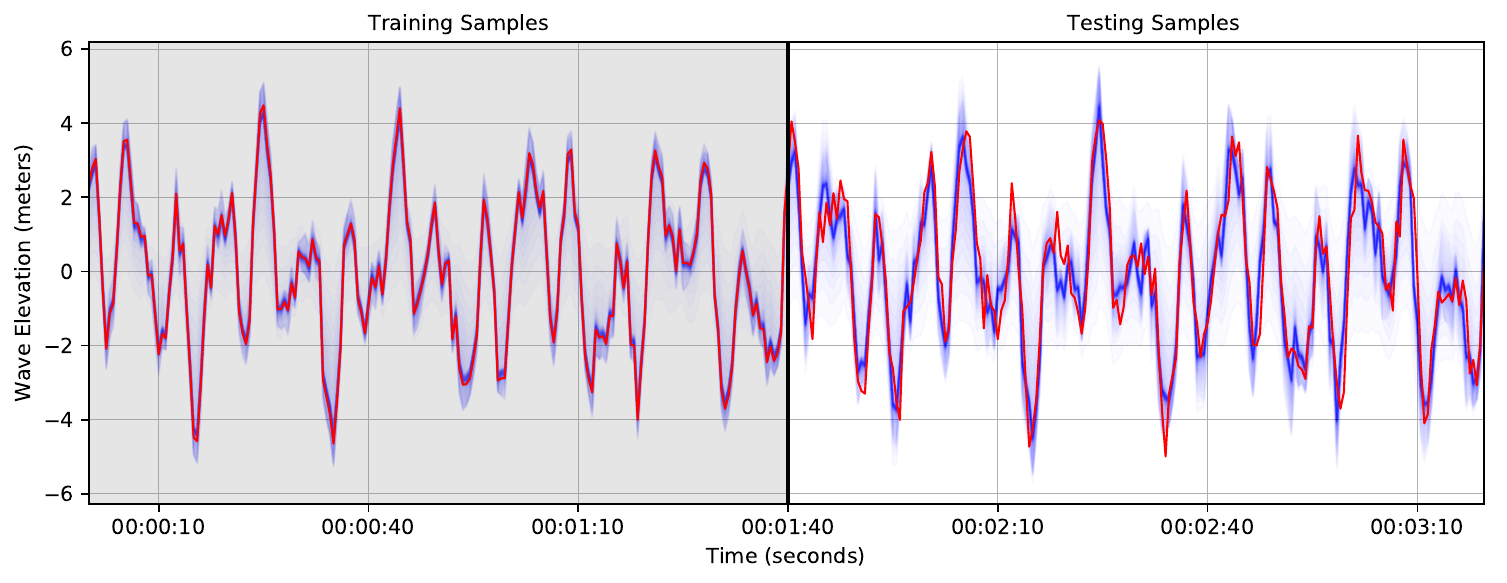}
	\end{center}
	\caption{Probabilistic forecasting of 50 prediction intervals for the simulated Ocean Wave Elevation time series.}
	\label{wave_PIs}
\end{figure}

\begin{figure}[h]
	\centering
	\includegraphics[width=1 \textwidth]{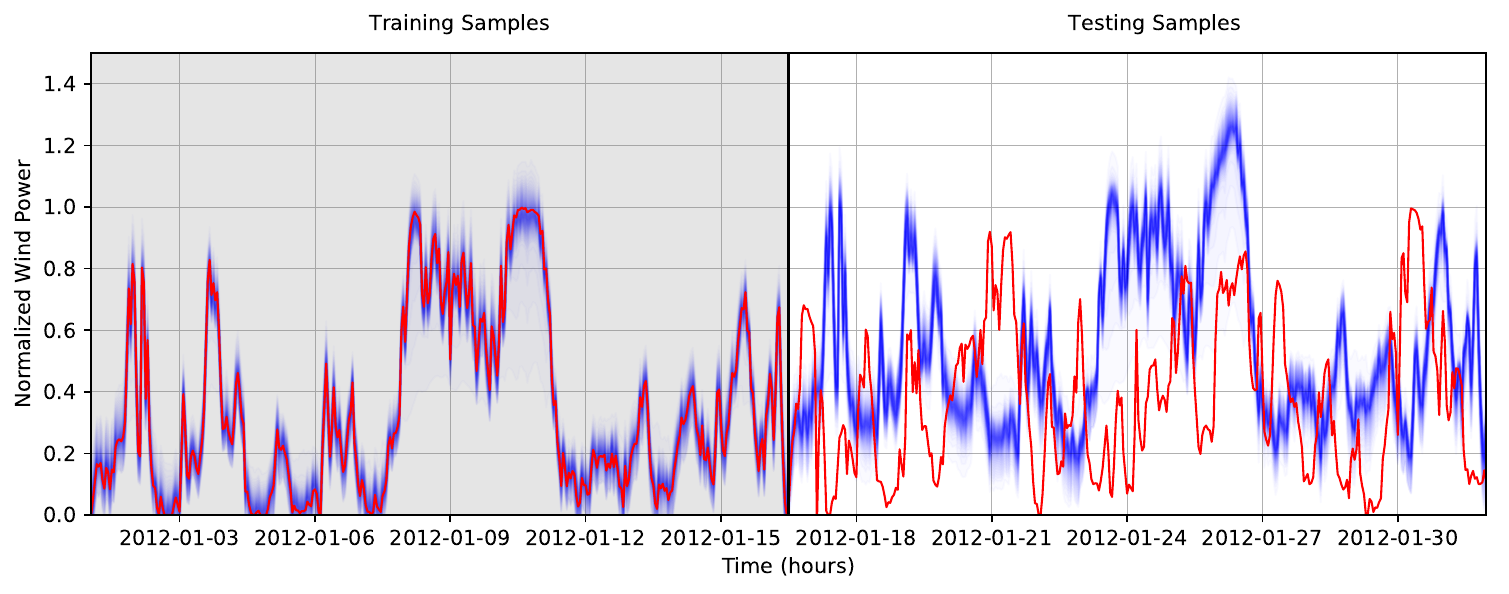}
	\caption{Probabilistic forecasting of 50 prediction intervals for the wind power time series.}
	\label{wind_PIs}
\end{figure}

\begin{figure}[h]
	\centering
	\includegraphics[width=1 \textwidth]{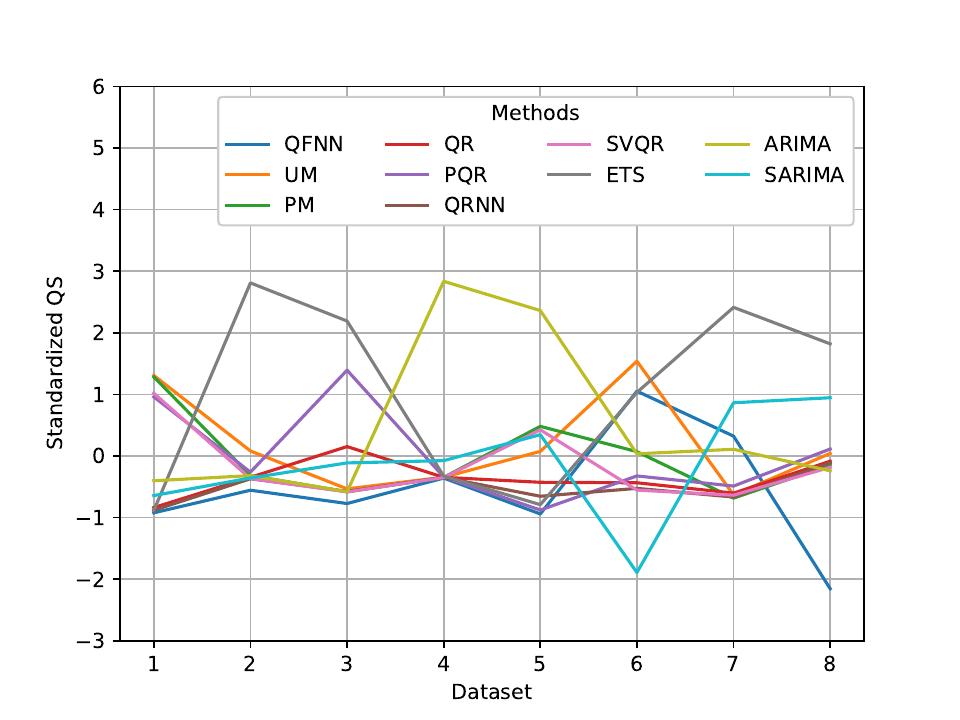}
	\caption{Standardized QS for QFNN and benchmark methods across the 8 datasets for estimating 100 quantiles. QFNN yields the lowest scores for each dataset except for solar and wind power where ETS and QR had the lowest scores.}
	\label{pi_QS}
\end{figure}

To establish QFNN's ability to estimate composite quantiles for multi-step probabilistic forecasting we conduct an experiment to predict 100 percentiles with nominal coverage rates equally space between 0 and 1. These quantile can be combined together to form 50 PIs with equally spaced PINC. We use QS to evaluate the 100 percentiles, and ACE and SS to evaluate the 50 PIs. Figures \ref{air_PIs} to \ref{wind_PIs} showcase QFNN estimation of 50 prediction intervals across test and training data from all 8 case studies. The red line in these figures represents the time series observations. Standardized QS for QFNN and benchmark methods across the 8 datasets for estimating 100 quantiles are presented in Fig. \ref{pi_QS}. We showcase standardized scores so as to make comparisons between datasets. 

\begin{figure}[h]
	\centering
	\includegraphics[width=1 \textwidth]{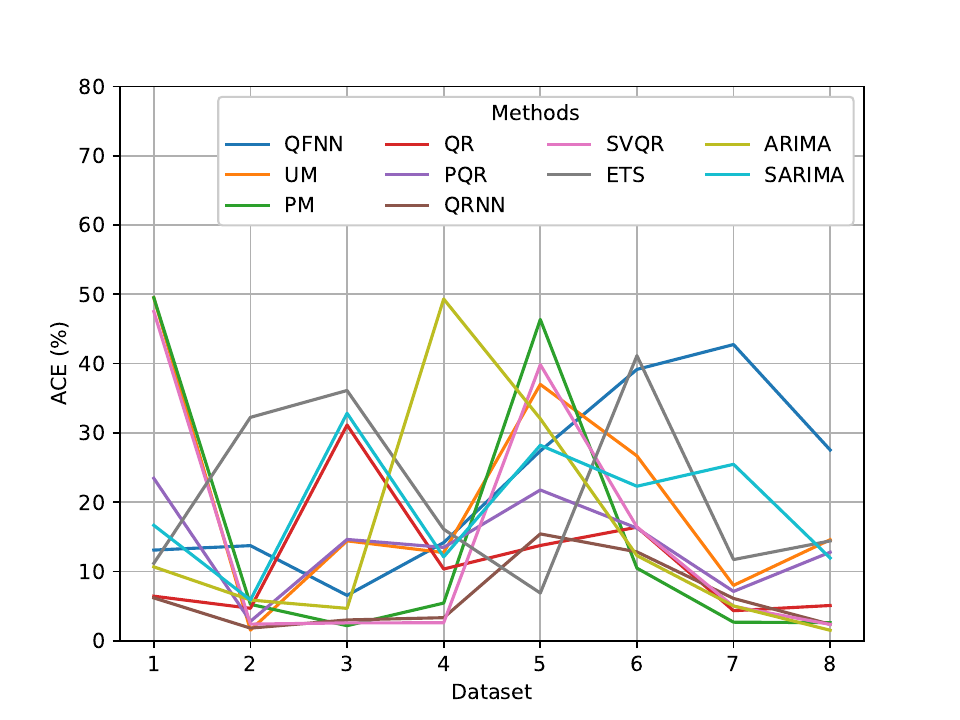}
	\caption{ACE scores for QFNN and benchmark methods across the 8 datasets for estimating 50 prediction intervals. QFNN averaged a 20\% coverage across the 8 datasets, and all methods yielded a high ACE for the stock price data and the lowest ACE across all methods was the wave elevation time series.}
	\label{pi_ACE}
\end{figure}

\begin{figure}[h]
	\centering
	\includegraphics[width=1 \textwidth]{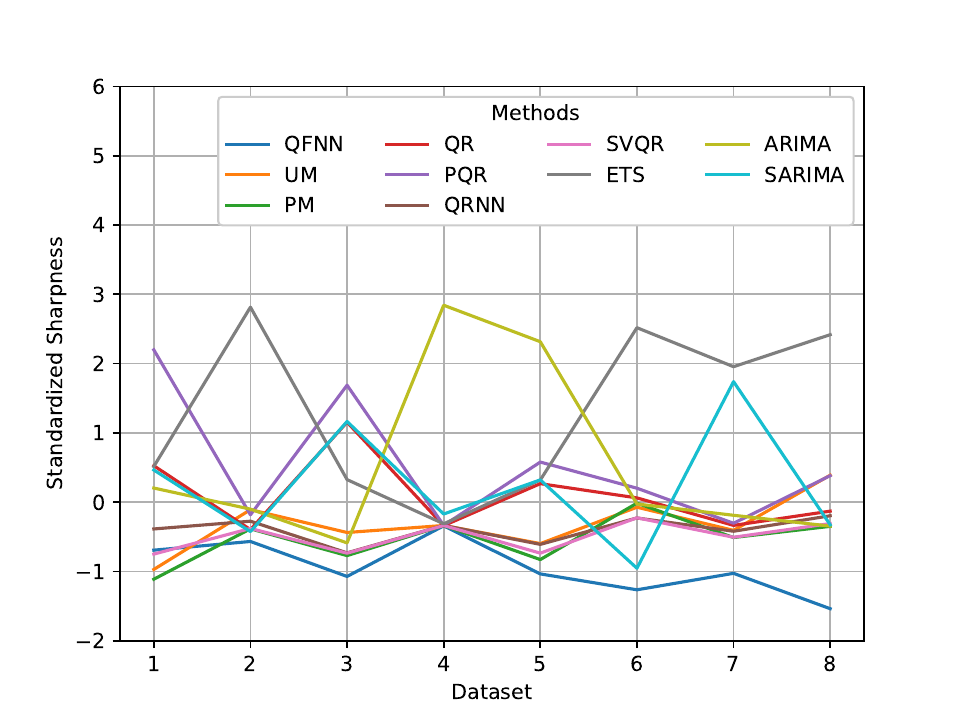}
	\caption{Standardized SS (sharpness) for QFNN and benchmark methods across the 8 datasets for estimating 50 prediction intervals. QFNN yields the lowest scores for each dataset except for the Air Passengers time series where the persistance method had the narrowest mean intervals.}
	\label{pi_sharp}
\end{figure}

QFNN yields the lowest scores for each dataset except for solar and wind power where ETS and PM had the lowest scores. ACE scores are presented in Fig. \ref{pi_ACE} in the form of a reliability plot. QFNN averaged a 23\% ACE rate across the 8 datasets, and most methods yielded a high ACE for the stock price data indicating poor coverage. The lowest ACE across all methods was the wave elevation time series. ARIMA had the best coverage for the wave elevation time series. Standardized SS is presented in Fig. \ref{pi_sharp} as a sharpness plot. QFNN yields the lowest sharpness for every dataset except for the Air Passengers time series where the persistence method had the narrowest mean intervals due to its variance estimation from the training data. The narrow QFNN intervals explain the poor coverage highlighted by the ACE scores.

\begin{figure}[h]
	\centering
	\includegraphics[width=1 \textwidth]{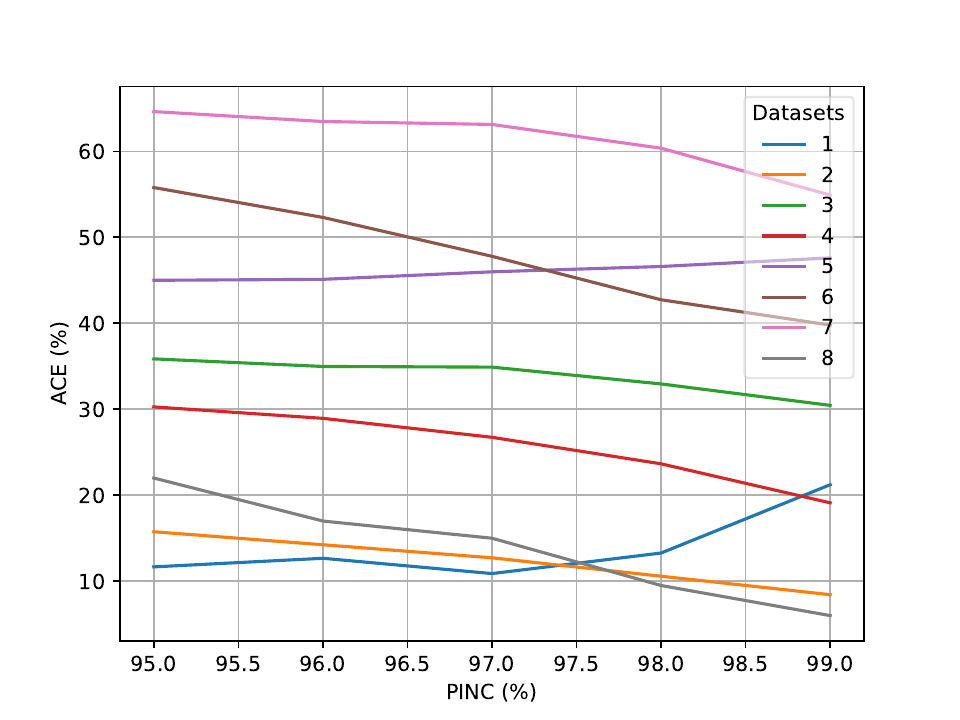}
	\caption{ACE for QFNN across the 8 datasets for estimating extreme valued PIs. The air passengers, sunspots, and wave elevation time series showed the best coverage of observations by the PINC rates, while wind power, solar power, and stock price time series resulted in poorer fits with high ACE scores up to 60\% for wind power.}
	\label{extreme_ACE}
\end{figure}

\begin{figure}[h]
	\centering
	\includegraphics[width=1 \textwidth]{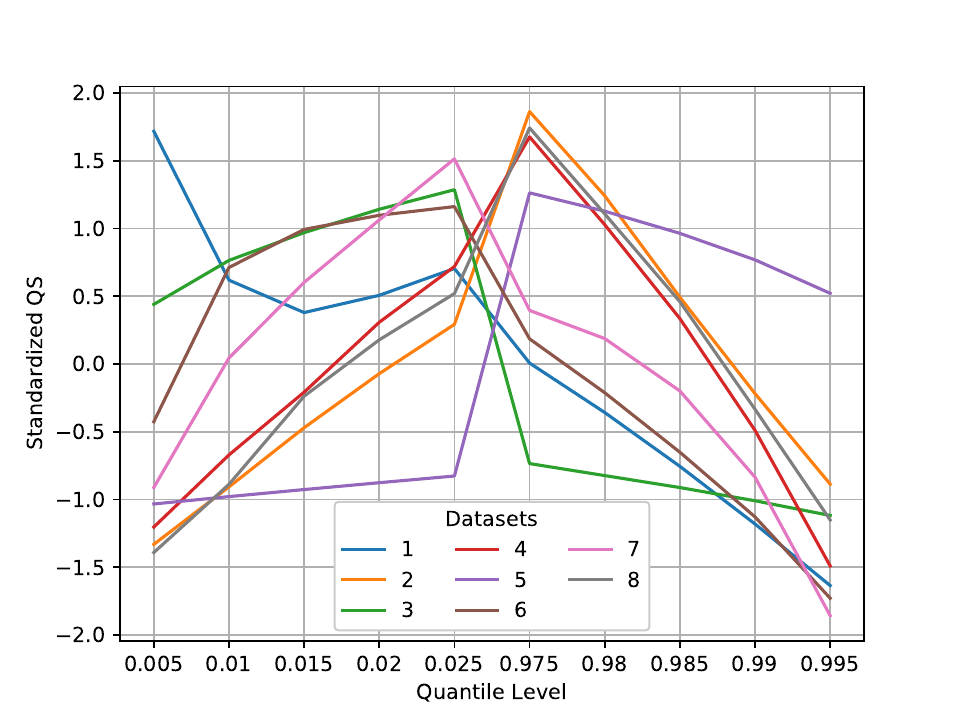}
	\caption{QS scores for QFNN across the 8 datasets for estimating extreme valued percentiles. QS for most time series was lower for extreme quantiles such as 0.5\% or 99.5\%, and higher for percentiles with coverage of  2.5\% and 97.5\%. 	}
	\label{extreme_QS}
\end{figure}

\begin{figure}[h]
	\centering
	\includegraphics[width=1 \textwidth]{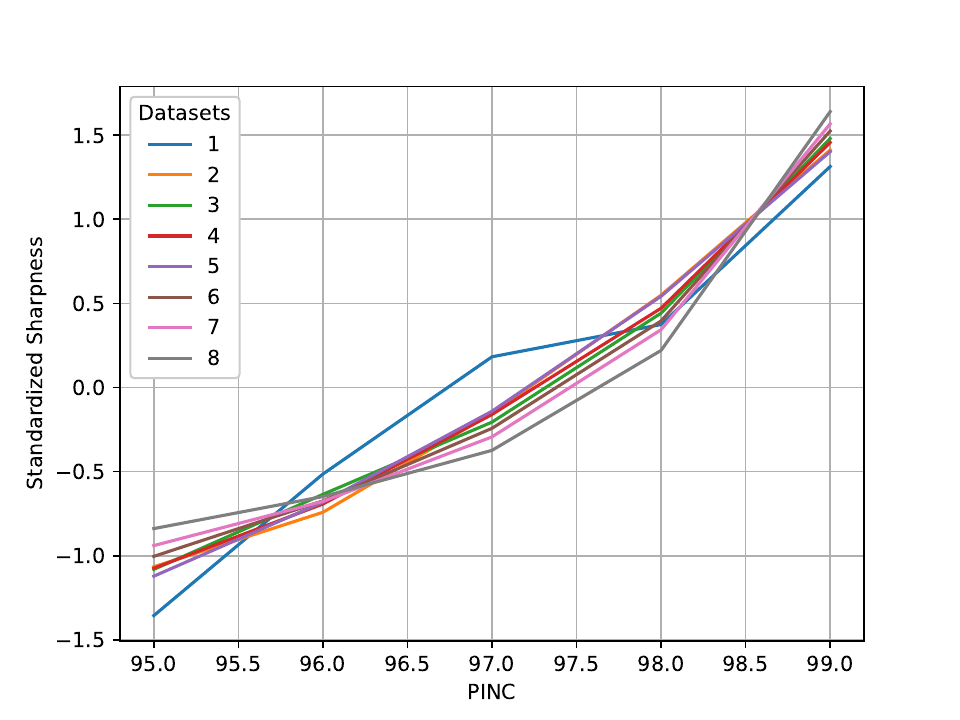}
	\caption{Standardized SS (sharpness) for QFNN across the 8 datasets for estimating extreme valued PIs. As the PINC increases, the width of the PIs increases.}
	\label{extreme_sharp}
\end{figure}

In many cases we are interested in extreme percentile values such as in anomaly detection or risk management in finance. Having well estimated extreme valued quantiles can have a big impact in these fields. Therefore, as over-viewed in the beginning of this section, we analyze ten low and high valued quantiles for QFNN, which combined form 5 high PINC values. As with multiple quantile evaluation above, we utilize QS to measure the performance of extreme percentiles, and ACE and SS to evaluate the PIs. In Fig. \ref{extreme_QS} QS plot is presented for QFNN across the 8 datasets for estimating extreme valued percentiles. QS for most time series is lower for extreme nominal rates such as 0.5\% or 99.5\%, and higher for percentiles with coverage of 2.5\% and 97.5\%. For PIs, we see in Fig. \ref{extreme_ACE} the reliability plot for QFNN. The air passengers, sunspots, and wave elevation time series showed the best coverage of observations by the PINC rates, while wind power, solar power, and stock price time series resulted in poorer fits with high ACE scores up to 64\% for wind power. Lastly, a sharpness plot for QFNN is presented in Fig. \ref{extreme_sharp}. As expected, as the PINC increases for PIs, the their width increases.

\subsubsection{Individual Case Study Analysis} \label{s:individual}

In our first individual case study analysis, we inspect the air passengers dataset, where the first six years of data (72 samples) are used for training QFNN and each benchmark method. The next six years were used for prediction. An ACF evaluation finds that the time series has a season of $ S = 12 $, and grid search found ARIMA(2,1,3) and SARIMA(1,0,0)(1,0,1)[12] to be the best hyperparameters for ARIMA and SARIMA respectively. We see in Fig. \ref{air_median} QRNN estimates the trend but not the seasonality. ETS and SARIMA estimate both trend and seasonality well, but the median forecasts fall below the test data. QFNN learns the shape of the data to capture the median appropriately. Fig. \ref{air_PIs} shows prediction intervals fitting the test data well. In the last 3 seanal periods, QFNN however does estimate percentiles above the observed data. For the remaining experiments, ARIMA and SARIMA hyperparameters are displayed in Table \ref{t:arima}.

Our second experiment demonstrates the power of QFNN in modeling non-constant seasonal patterns. The sunspots case study is used which has seasonal patterns of varying amplitudes. The years from 1700 to 1858 were used for training, and the years 1859 to 2017 were used for testing. We see in Fig. \ref{sun_median} that QRNN fails to capture any meaningful pattern in its prediction. SARIMA captures a seasonal pattern that is out of phase with the sunspot test series, and ETS shots off in the test set with a positive trend. QFNN captures a seasonal pattern that is a bit more in phase with the number of sunspots over the years and is also able to learn multiple patterns of the sunspot time series. Fig. \ref{sun_PIs} shows prediction intervals fitting the test data surprising well, QFNN captures higher peaks around 1943 and lower peaks around 1903. 

The third experiment uses the real-time load demand case study. We use the first 372 hours in the time series for training. In the median plot of Fig. \ref{load_median} SARIMA captures the daily seasonality but fails to capture any cycles in the test set, and ETS goes up in the test set with a positive trend. QFNN learns both the seasonal and cyclical pattern of the load demand. The capture of the cyclical pattern in the load data by QFNN is better presented in Fig. \ref{load_PIs} where we see a tight fit of the daily seasonal and weekly cyclical pattern. The only deviation being around January 21, 2017, which shows a lower observed demand in load possibly due to a warmer weekend and less power needed for heating. The fourth experiment uses the solar power case study and uses the first 380 hours of training. The training set includes samples from both sunny and non-sunny days where solar power is lower than average. QFNN learns a constant daily quantile pattern for all its prediction intervals in Fig. \ref{solar_PIs}. 

The fifth experiment using the closing stock prices of the Apple corporation is considered a fascinating case study due to the highly random nature of stock movements. For training, the first 2.5 years of closing prices are used, and testing is composed of closing prices up until the start of April 2018. In Fig. \ref{stock_median} we see a long-term positive trend and possibly a cycle in the stock price of Apple across the five years. In the plot, SVQR fits the linear trend of the stock series but nothing else. ETS learns a non-existing seasonal pattern, and SARIMA doesn't seem to capture any meaningful pattern. With QFNN we see that it learns the cyclic and positive trend of the stock price which follows the test data better than any other method. This is also demonstrated in Fig. \ref{stock_PIs} that up until the end of 2017 QFNN follows the trend and cycle, but then in 2018, the price of Apple jumps higher than the prediction. 

The prediction intervals forecasted by QFNN for the simulated wave elevation case study is shown in Fig. \ref{wave_PIs} and the median predictions in Fig. \ref{wave_median}. It is not a surprise that since ocean waves can be modeled by a sum of sinusoids that QFNN can estimate well the amplitudes, frequencies, and phases of the irregular periodic patterns. For the median percentile predictions, QRNN captures no meaningful pattern, and ETS and SARIMA both capture a periodic pattern whose phase and frequency do not visually match the observed wave elevations.

The remaining experiments are the internet and wind case studies. Median plots of these experiments are not shown since the benchmark methods performed poorly in capturing the multiple or irregular seasonal patterns in these time series. We present the prediction intervals by QFNN in Fig. \ref{internet_PIs} where the first 343 hours are used for training. We see that QFNN can learn the multiple seasonal patterns of internet traffic data. The last experiment conducted is on the wind power dataset. In Fig. \ref{wind_PIs} we see no identifiable periodic or aperiodic patterns in the wind time series training set. This explains why QFNN has a hard time forecasting the test set. We do see a few peaks predicted by QFNN such as on January 27 and the 30th, but overall the PF is very poor. The wind experiment demonstrates that nonstationary time series without periodic patterns or trend can not be predicted by QFNN. 

\section{Conclusion} \label{s:con}

Probabilistic predictions can provide a much better analysis of uncertainty then point forecasts. In this paper, a novel approach for probabilistic forecasting is presented called the quantile Fourier neural network. The proposed approach uses a smooth approximation to the pinball ball loss function for estimating composite quantiles. Furthermore, the proposed model provides forecasts using extrapolation based regression instead of autoregression. Extrapolation based regression has not been studied before for probabilistic forecasting. Empirical results on real world univariate time series showcase that our model is able to appropriately capture periodic and aperiodic components to provide high-quality probabilistic predictions. 

From the thorough skill, reliability, and sharpness analysis and visual inspection of median percentile, 100 percentile, and extreme value percentile predictions we can form the following conclusions on which circumstances QFNN performs well and why. QFNN is able to appropriately capture the phase, frequency, and amplitude of periodic components of seasonal time series. This is evident in its quantile estimates for seasonal time series such as wave elevation and load demand. QFNN is also able to capture multiple seasonality such as in the fit for the time series of sunspots and internet traffic. Lastly, QFNN performs well when linear trend is present such as in the air passengers time series, and is even able to loosely capture long term cycles such as in the stock price time series. However, the shortcomings of QFNN are that it is unable to capture patterns in more chaotic data such as wind power and short term stock prices. Another shortcoming is that QFNN is designed for signal extrapolation, and thus it will repeat the same patterns it learned during training on testing data. This can make it ideal for time series imputation. But it is unable to predict unforeseen patterns or patterns dependent on other variables beyond time. For instance, QFNN is not able to predict non-sunny days for solar power, it performs bad on wind power, and it is not able to predict the daily values of stock prices. 

Given the novelty of our approach, more research needs to be conducted to assess its application to more domains and under different scenarios. Further studies could also look at other cost functions such as using the interval score that measures overall skill of PIs and sharpness \cite{gneiting2007strictly}. This could provide appropriate prediction intervals that may have even higher reliability and while maintaining narrow intervals. Furthermore, the influence of exogenous variables as additional inputs to the quantile Fourier neural network could be explored.

\section*{References}

\bibliographystyle{elsarticle-num}
\bibliography{mybib}

\begin{thebibliography}{10}
\expandafter\ifx\csname url\endcsname\relax
  \def\url#1{\texttt{#1}}\fi
\expandafter\ifx\csname urlprefix\endcsname\relax\def\urlprefix{URL }\fi
\expandafter\ifx\csname href\endcsname\relax
  \def\href#1#2{#2} \def\path#1{#1}\fi

\bibitem{xue2017financial}
J.~Xue, S.~Zhou, Q.~Liu, X.~Liu, J.~Yin, Financial time series prediction using
  ℓ2, 1rf-elm, Neurocomputing.

\bibitem{hu2016predicting}
Y.~Hu, C.~Hu, S.~Fu, P.~Shi, B.~Ning, Predicting the popularity of viral topics
  based on time series forecasting, Neurocomputing 210 (2016) 55--65.

\bibitem{zhou2017delta}
T.~Zhou, G.~Han, X.~Xu, Z.~Lin, C.~Han, Y.~Huang, J.~Qin, $\delta$-agree
  adaboost stacked autoencoder for short-term traffic flow forecasting,
  Neurocomputing 247 (2017) 31--38.

\bibitem{yan2016time}
J.~Yan, K.~Li, E.~Bai, Z.~Yang, A.~Foley, Time series wind power forecasting
  based on variant gaussian process and tlbo, Neurocomputing 189 (2016)
  135--144.

\bibitem{godfrey2017neural}
L.~B. Godfrey, M.~S. Gashler, Neural decomposition of time-series data for
  effective generalization, arXiv preprint arXiv:1705.09137.

\bibitem{box2015time}
G.~E. Box, G.~M. Jenkins, G.~C. Reinsel, G.~M. Ljung, Time series analysis:
  forecasting and control, John Wiley \& Sons, 2015.

\bibitem{langkvist2014review}
M.~L{\"a}ngkvist, L.~Karlsson, A.~Loutfi, A review of unsupervised feature
  learning and deep learning for time-series modeling, Pattern Recognition
  Letters 42 (2014) 11--24.

\bibitem{doherty2005new}
R.~Doherty, M.~O'Malley, A new approach to quantify reserve demand in systems
  with significant installed wind capacity, IEEE Transactions on Power Systems
  20~(2) (2005) 587--595.

\bibitem{castronuovo2004optimization}
E.~D. Castronuovo, J.~P. Lopes, On the optimization of the daily operation of a
  wind-hydro power plant, IEEE Transactions on Power Systems 19~(3) (2004)
  1599--1606.

\bibitem{pinson2007trading}
P.~Pinson, C.~Chevallier, G.~N. Kariniotakis, Trading wind generation from
  short-term probabilistic forecasts of wind power, IEEE Transactions on Power
  Systems 22~(3) (2007) 1148--1156.

\bibitem{yang2017power}
Y.~Yang, S.~Li, W.~Li, M.~Qu, Power load probability density forecasting using
  gaussian process quantile regression, Applied Energy.

\bibitem{taieb2016forecasting}
S.~B. Taieb, R.~Huser, R.~J. Hyndman, M.~G. Genton, Forecasting uncertainty in
  electricity smart meter data by boosting additive quantile regression, IEEE
  Transactions on Smart Grid 7~(5) (2016) 2448--2455.

\bibitem{su2014stochastic}
W.~Su, J.~Wang, J.~Roh, Stochastic energy scheduling in microgrids with
  intermittent renewable energy resources, IEEE Transactions on Smart Grid
  5~(4) (2014) 1876--1883.

\bibitem{hong2014long}
T.~Hong, J.~Wilson, J.~Xie, Long term probabilistic load forecasting and
  normalization with hourly information, IEEE Transactions on Smart Grid 5~(1)
  (2014) 456--462.

\bibitem{botterud2013demand}
A.~Botterud, Z.~Zhou, J.~Wang, J.~Sumaili, H.~Keko, J.~Mendes, R.~J. Bessa,
  V.~Miranda, Demand dispatch and probabilistic wind power forecasting in unit
  commitment and economic dispatch: A case study of illinois, IEEE Transactions
  on Sustainable Energy 4~(1) (2013) 250--261.

\bibitem{andrade2017probabilistic}
J.~R. Andrade, J.~Filipe, M.~Reis, R.~J. Bessa, Probabilistic price forecasting
  for day-ahead and intraday markets: Beyond the statistical model,
  Sustainability 9~(11) (2017) 1990.

\bibitem{zhang2014review}
Y.~Zhang, J.~Wang, X.~Wang, Review on probabilistic forecasting of wind power
  generation, Renewable and Sustainable Energy Reviews 32 (2014) 255--270.

\bibitem{van2017review}
D.~van~der Meer, J.~Wid{\'e}n, J.~Munkhammar, Review on probabilistic
  forecasting of photovoltaic power production and electricity consumption,
  Renewable and Sustainable Energy Reviews.

\bibitem{hong2016probabilistic}
T.~Hong, P.~Pinson, S.~Fan, H.~Zareipour, A.~Troccoli, R.~J. Hyndman,
  Probabilistic energy forecasting: Global energy forecasting competition 2014
  and beyond (2016).

\bibitem{makridakis2018m4}
S.~Makridakis, E.~Spiliotis, V.~Assimakopoulos, The m4 competition: Results,
  findings, conclusion and way forward, International Journal of Forecasting.

\bibitem{huang2017deterministic}
C.-M. Huang, Y.-C. Huang, K.-Y. Huang, S.-J. Chen, S.-P. Yang, Deterministic
  and probabilistic wind power forecasting using a hybrid method, in:
  Industrial Technology (ICIT), 2017 IEEE International Conference on, IEEE,
  2017, pp. 400--405.

\bibitem{yang2015deep}
J.~Yang, M.~N. Nguyen, P.~P. San, X.~Li, S.~Krishnaswamy, Deep convolutional
  neural networks on multichannel time series for human activity recognition.,
  in: Ijcai, Vol.~15, 2015, pp. 3995--4001.

\bibitem{gashler2016modeling}
M.~S. Gashler, S.~C. Ashmore, Modeling time series data with deep fourier
  neural networks, Neurocomputing 188 (2016) 3--11.

\bibitem{germec2009fourier}
K.~E. Germec, Fourier neural networks for real-time harmonic analysis, in:
  Signal Processing and Communications Applications Conference, 2009. SIU 2009.
  IEEE 17th, IEEE, 2009, pp. 333--336.

\bibitem{pinson2007non}
P.~Pinson, H.~A. Nielsen, J.~K. M{\o}ller, H.~Madsen, G.~N. Kariniotakis,
  Non-parametric probabilistic forecasts of wind power: required properties and
  evaluation, Wind Energy 10~(6) (2007) 497--516.

\bibitem{koenker1978regression}
R.~Koenker, G.~Bassett~Jr, Regression quantiles, Econometrica: journal of the
  Econometric Society (1978) 33--50.

\bibitem{koenker2005quantile}
R.~Koenker, Quantile regression, no.~38, Cambridge university press, 2005.

\bibitem{bremnes2004probabilistic}
J.~B. Bremnes, Probabilistic wind power forecasts using local quantile
  regression, Wind Energy 7~(1) (2004) 47--54.

\bibitem{nielsen2006using}
H.~A. Nielsen, H.~Madsen, T.~S. Nielsen, Using quantile regression to extend an
  existing wind power forecasting system with probabilistic forecasts, Wind
  Energy 9~(1-2) (2006) 95--108.

\bibitem{landry2016probabilistic}
M.~Landry, T.~P. Erlinger, D.~Patschke, C.~Varrichio, Probabilistic gradient
  boosting machines for gefcom2014 wind forecasting, International Journal of
  Forecasting 32~(3) (2016) 1061--1066.

\bibitem{juban2016multiple}
R.~Juban, H.~Ohlsson, M.~Maasoumy, L.~Poirier, J.~Z. Kolter, A multiple
  quantile regression approach to the wind, solar, and price tracks of
  gefcom2014, International Journal of Forecasting 32~(3) (2016) 1094--1102.

\bibitem{juban2008uncertainty}
J.~Juban, L.~Fugon, G.~Kariniotakis, Uncertainty estimation of wind power
  forecasts: Comparison of probabilistic modelling approaches, in: European
  Wind Energy Conference \& Exhibition EWEC 2008, EWEC, 2008, pp. 10--pages.

\bibitem{taylor2000quantile}
J.~W. Taylor, A quantile regression neural network approach to estimating the
  conditional density of multiperiod returns, Journal of Forecasting 19~(4)
  (2000) 299--311.

\bibitem{xu2016quantile}
Q.~Xu, X.~Liu, C.~Jiang, K.~Yu, Quantile autoregression neural network model
  with applications to evaluating value at risk, Applied Soft Computing 49
  (2016) 1--12.

\bibitem{cannon2011quantile}
A.~J. Cannon, Quantile regression neural networks: Implementation in r and
  application to precipitation downscaling, Computers \& Geosciences 37~(9)
  (2011) 1277--1284.

\bibitem{grushka2016quantile}
Y.~Grushka-Cockayne, K.~C. Lichtendahl, V.~R.~R. Jose, R.~L. Winkler, Quantile
  evaluation, sensitivity to bracketing, and sharing business payoffs.

\bibitem{parascandolo2016taming}
G.~Parascandolo, H.~Huttunen, T.~Virtanen, Taming the waves: sine as activation
  function in deep neural networks.

\bibitem{silvescu1997new}
A.~Silvescu, A new kind of neural networks (1997).

\bibitem{silvescu1999fourier}
A.~Silvescu, Fourier neural networks, in: Neural Networks, 1999. IJCNN'99.
  International Joint Conference on, Vol.~1, IEEE, 1999, pp. 488--491.

\bibitem{minami1999real}
K.-i. Minami, H.~Nakajima, T.~Toyoshima, Real-time discrimination of
  ventricular tachyarrhythmia with fourier-transform neural network, IEEE
  transactions on Biomedical Engineering 46~(2) (1999) 179--185.

\bibitem{mingo2004fourier}
L.~Mingo, L.~Aslanyan, J.~Castellanos, M.~Diaz, V.~Riazanov, Fourier neural
  networks: An approach with sinusoidal activation functions.

\bibitem{tan2006fourier}
H.~Tan, Fourier neural networks and generalized single hidden layer networks in
  aircraft engine fault diagnostics, Journal of engineering for gas turbines
  and power 128~(4) (2006) 773--782.

\bibitem{zuo2009fourier}
W.~Zuo, Y.~Zhu, L.~Cai, Fourier-neural-network-based learning control for a
  class of nonlinear systems with flexible components, IEEE transactions on
  neural networks 20~(1) (2009) 139--151.

\bibitem{tan2017use}
D.~Tan, W.~Chen, H.~Wang, On the use of monte-carlo simulation and deep fourier
  neural network in lane departure warning, IEEE Intelligent Transportation
  Systems Magazine 9~(4) (2017) 76--90.

\bibitem{gashler2014training}
M.~S. Gashler, S.~C. Ashmore, Training deep fourier neural networks to fit
  time-series data, in: International Conference on Intelligent Computing,
  Springer, 2014, pp. 48--55.

\bibitem{zheng2011gradient}
S.~Zheng, Gradient descent algorithms for quantile regression with smooth
  approximation, International Journal of Machine Learning and Cybernetics
  2~(3) (2011) 191.

\bibitem{srivastava2014dropout}
N.~Srivastava, G.~Hinton, A.~Krizhevsky, I.~Sutskever, R.~Salakhutdinov,
  Dropout: a simple way to prevent neural networks from overfitting, The
  Journal of Machine Learning Research 15~(1) (2014) 1929--1958.

\bibitem{brockwell2002introduction}
P.~J. Brockwell, R.~A. Davis, M.~V. Calder, Introduction to time series and
  forecasting, Vol.~2, Springer, 2002.

\bibitem{chollet2015keras}
F.~Chollet, et~al., Keras: Deep learning library for theano and tensorflow,
  URL: https://keras. io/k 7~(8).

\bibitem{air}
Box, Jenkins,
  \href{http://datamarket.com/data/list/?q=provider:tsdl}{International airline
  passengers: monthly totals in thousands. jan 49 to dec 60}, Time Series Data
  Library(Date last accessed 5-April-2018).
\newline\urlprefix\url{http://datamarket.com/data/list/?q=provider:tsdl}

\bibitem{sidc}
S.~W.~D. Center, \href{http://www.sidc.be/silso/datafiles}{Sunspot data from
  the world data center silso, royal observatory of belgium, brussels},
  International Sunspot Number Monthly Bulletin and online catalogue(Date last
  accessed 5-April-2018).
\newline\urlprefix\url{http://www.sidc.be/silso/datafiles}

\bibitem{load}
\href{https://www.iso-ne.com/isoexpress/web/reports/load-and-demand/-/tree/zone-info}{Iso
  ne ca, rt demand, from 1-jan-2017 to 1-feb-2017}, ISO New England Public(Date
  last accessed 5-April-2018).
\newline\urlprefix\url{https://www.iso-ne.com/isoexpress/web/reports/load-and-demand/-/tree/zone-info}

\bibitem{internet}
\href{https://datamarket.com/data/list/?q=provider:tsdl}{Internet traffic data
  (in bits) from a private isp with centres in 11 european cities. from
  18-june-2005 to 16-july-2005}, Time Series Data Library(Date last accessed
  5-April-2018).
\newline\urlprefix\url{https://datamarket.com/data/list/?q=provider:tsdl}

\bibitem{fama1988business}
E.~F. Fama, K.~R. French, Business cycles and the behavior of metals prices,
  The Journal of Finance 43~(5) (1988) 1075--1093.

\bibitem{aapl}
\href{https://finance.yahoo.com/quote/AAPL?p=AAPL}{Apple inc. (aapl) closing
  stock price. from 3-january-2012 to 13-april-2018}, Yahoo Finance(Date last
  accessed 13-April-2018).
\newline\urlprefix\url{https://finance.yahoo.com/quote/AAPL?p=AAPL}

\bibitem{falnes2002ocean}
J.~Falnes, Ocean waves and oscillating systems: linear interactions including
  wave-energy extraction, Cambridge university press, 2002.

\bibitem{sideratos2012probabilistic}
G.~Sideratos, N.~D. Hatziargyriou, Probabilistic wind power forecasting using
  radial basis function neural networks, IEEE Transactions on Power Systems
  27~(4) (2012) 1788--1796.

\bibitem{wan2014optimal}
C.~Wan, Z.~Xu, P.~Pinson, Z.~Y. Dong, K.~P. Wong, Optimal prediction intervals
  of wind power generation, IEEE Transactions on Power Systems 29~(3) (2014)
  1166--1174.

\bibitem{arima}
Generating quantile forecasts in r,
  \url{https://robjhyndman.com/hyndsight/quantile-forecasts-in-r/}, accessed:
  20-April-2018.

\bibitem{hyndman2002state}
R.~J. Hyndman, A.~B. Koehler, R.~D. Snyder, S.~Grose, A state space framework
  for automatic forecasting using exponential smoothing methods, International
  Journal of forecasting 18~(3) (2002) 439--454.

\bibitem{sarimax}
Seasonal autoregressive integrated moving average with exogenous regressors
  model in python,
  \url{http://www.statsmodels.org/dev/generated/statsmodels.tsa.statespace.sarimax.SARIMAX.html},
  accessed: 20-April-2018.

\bibitem{holt}
Python implementation of holt-winters seasonal methods,
  \url{https://gist.github.com/andrequeiroz/5888967}, accessed: 20-April-2018.

\bibitem{hatalis2018empirical}
K.~Hatalis, S.~Kishore, K.~Scheinberg, A.~Lamadrid, An empirical analysis of
  constrained support vector quantile regression for nonparametric
  probabilistic forecasting of wind power, arXiv preprint arXiv:1803.10888.

\bibitem{hatalis2017smooth}
K.~Hatalis, A.~J. Lamadrid, K.~Scheinberg, S.~Kishore, Smooth pinball neural
  network for probabilistic forecasting of wind power, arXiv preprint
  arXiv:1710.01720.

\bibitem{steinwart2017liquidsvm}
I.~Steinwart, P.~Thomann, liquidsvm: A fast and versatile svm package, arXiv
  preprint arXiv:1702.06899.

\bibitem{gneiting2007strictly}
T.~Gneiting, A.~E. Raftery, Strictly proper scoring rules, prediction, and
  estimation, Journal of the American Statistical Association 102~(477) (2007)
  359--378.

\end{thebibliography}

\newpage

\newpage

\begin{sidewaystable}
	\centering
	\caption{Datasets used in the experiments.}
	\label{t:cases}
	\begin{tabular*}{\textwidth}{@{\extracolsep{\fill}}clccc@{}}
		\toprule
		Case Study & Target & Samples & Time Granularity & Reference \\
		\midrule
		1 & Air Passengers & 144 & Month & \cite{air} \\
		2 & Sunspots & 318 & Year & \cite{sidc} \\
		3 & Real-Time Load Demand & 744 & Hour & \cite{load} \\
		4 & Internet Traffic Data (in bits) & 686 & Hour & \cite{internet} \\
		5 & Apple Closing Stock Price & 1581 & Day & \cite{aapl} \\
		6 & Solar Power & 760 & Hour & \cite{hong2016probabilistic} \\
		7 & Wind Power & 744 & Hour & \cite{hong2016probabilistic} \\
		8 & Ocean Wave Elevation & 400 & Second & (simulated) \\
		\bottomrule
	\end{tabular*}
\end{sidewaystable}

\newpage

\begin{table*}[t]
	\centering
	\caption{Hyperparameters estimated by grid search for ARIMA and SARIMA for each case study. The seasonal term S is estimated using the ACF plot.}
	\label{t:arima}
	\begin{tabular*}{12cm}{@{\extracolsep{\fill}}cccc@{}}
		\toprule
		Case Study & ARIMA(p,d,q) & SARIMA(p,d,q)(P,D,Q) & S \\
		\midrule
		1 & (2, 1, 3) & (1, 0, 0)(1, 0, 1) & 12 \\
		2 & (3, 1, 2) & (1, 0, 1)(0, 1, 1) & 10 \\
		3 & (2, 1, 3) & (1, 1, 1)(1, 1, 1) & 24 \\
		4 & (2, 2, 2) & (1, 1, 1)(1, 1, 1) & 24 \\
		5 & (2, 2, 2) & (1, 1, 1)(0, 1, 1) & 149 \\
		6 & (2, 1, 2) & (1, 0, 1)(1, 0, 1) & 24 \\
		7 & (2, 0, 1) & (1, 1, 1)(0, 0, 0) & 24 \\
		8 & (2, 0, 2) & (1, 0, 1)(0, 1, 1) & 39 \\
		\bottomrule
	\end{tabular*}
\end{table*}

\newpage

\newpage

\newpage

\end{document}